\documentclass[11pt]{article}

\usepackage[preprint]{acl}

\usepackage{times}
\usepackage{latexsym}

\usepackage[T1]{fontenc}
\usepackage{graphicx}
\graphicspath{{figs/}}
\usepackage{pifont}
\usepackage{adjustbox}
\usepackage{array}
\usepackage{booktabs}
\usepackage{arydshln}
\usepackage{booktabs}
\usepackage{multirow}

\usepackage[most]{tcolorbox}   

\usepackage{multicol}

\usepackage[utf8]{inputenc}

\usepackage{microtype}

\usepackage{inconsolata}

\newcommand{\cmark}{\ding{51}} 
\newcommand{\xmark}{\ding{55}} 
\usepackage{amsmath}   
\usepackage{amssymb}   

\usepackage{tabularx}  

\title{MathSight: A Benchmark Exploring Have Vision-Language Models Really Seen in University-Level Mathematical Reasoning?}

\author{
 \textbf{Yuandong Wang\textsuperscript{1}},
 \textbf{Yao Cui\textsuperscript{1}},
 \textbf{Yuxin Zhao\textsuperscript{1}},
 \textbf{Zhen Yang*\textsuperscript{2}},
 \textbf{Yangfu Zhu\textsuperscript{1}},
 \textbf{Zhenzhou Shao\textsuperscript{1}}
\\
 \textsuperscript{1}Capital Normal University,
 \textsuperscript{2}Tsinghua University
\\
 \small{
   \textbf{*Correspondence:} \href{mailto:email@domain}{yang-zhen@mail.tsinghua.edu.cn}
 }
 \\
  \small{
   \textbf{Project Page:} {https://cnu-bot-group.github.io/MathSight/}
 }
}

\begin{document}
\maketitle
\begin{abstract}



Recent advances in Vision-Language Models (VLMs) have achieved impressive progress in multimodal mathematical reasoning.
Yet, how much visual information truly contributes to reasoning remains unclear.
Existing benchmarks report strong overall performance but seldom isolate the role of the image modality, leaving open whether VLMs genuinely leverage visual understanding or merely depend on linguistic priors.
To address this, we present \textbf{MathSight}, a university-level multimodal mathematical reasoning benchmark designed to disentangle and quantify the effect of visual input.
Each problem includes multiple visual variants—original, hand-drawn, photo-captured—and a text-only condition for controlled comparison.
Experiments on state-of-the-art VLMs reveal a consistent trend: the contribution of visual information diminishes with increasing problem difficulty.
Remarkably, Qwen3-VL without any image input surpasses both its multimodal variants and GPT-5, underscoring the need for benchmarks like MathSight to advance genuine vision-grounded reasoning in future models.

\end{abstract}

\section{Introduction}
Recent advances in Vision-Language Models (VLMs) have led to significant progress in multimodal reasoning, particularly in mathematical problem solving. A growing collection of benchmarks, such as MathVista~\cite{lu2023mathvista}, MathVerse~\cite{zhang2025mathverse}, MATH-Vision~\cite{wang2024math-vision}, MathCheck~\cite{zhou2024your}, and Dynamath~\cite{Zou:24DynaMath}, has catalyzed rapid progress in this field, driving models toward increasingly sophisticated multimodal reasoning capabilities. 

\begin{figure*}[t!]
    \centering
    \includegraphics[width=\linewidth]{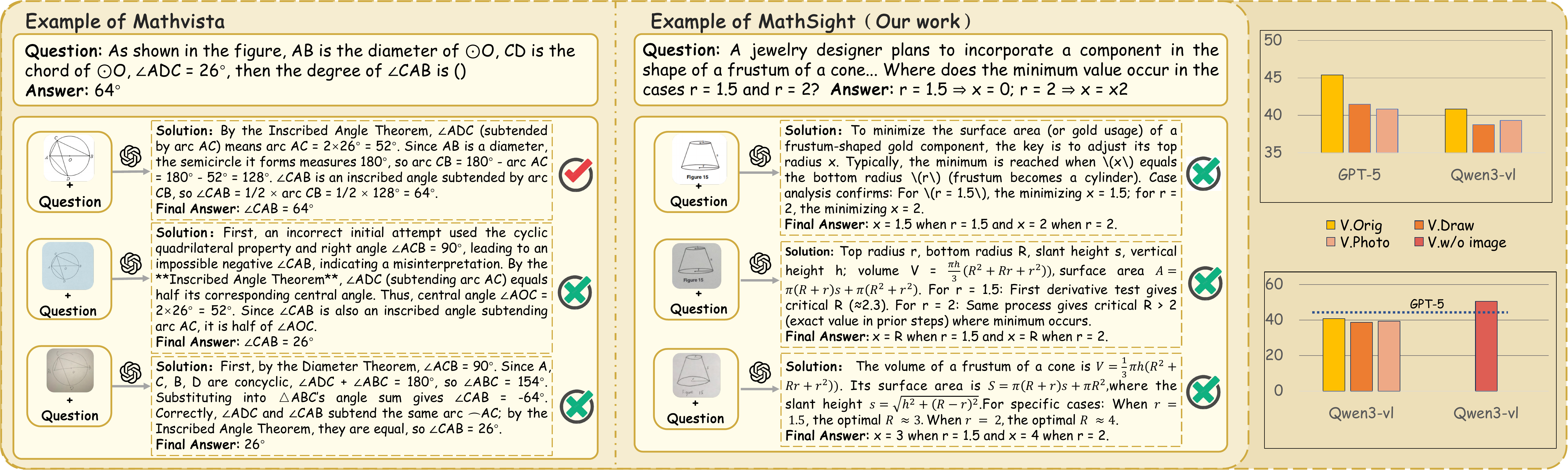}
    \caption{Motivation behind \textbf{MathSight}. Our main motivation is to investigate how different image versions affect the performance of VLMs on university-level and beyond mathematical reasoning problems. Since many existing studies focus on university-level (and below) math, as shown in Table 1, we deliberately curated university-level questions when constructing our dataset. }
    \label{fig:motivation_example}
\end{figure*}

Despite these advancements, the role of visual information in mathematical reasoning remains both underexplored and ambiguous. While existing benchmarks report impressive overall scores, they rarely isolate how much the image modality actually contributes to reasoning. As a result, it is still unclear whether current VLMs genuinely perform vision-grounded reasoning or merely rely on textual priors to approximate correct answers. As shown in Figure~\ref{fig:motivation_example}, existing multimodal mathematical reasoning benchmarks typically include only a single, clean visual representation for each problem. However, after intentionally introducing additional visual variants, including \textit{photo-captured} and \textit{hand-drawn} versions, we find that many problems that models previously answered correctly became incorrect. This observation reveals that even subtle changes in visual appearance can disrupt the reasoning process of current VLMs, suggesting that their ``visual understanding'' may be superficial and overly dependent on presentation style rather than underlying geometry or semantics. This motivates our systematic investigation of how visual variation affects reasoning stability and accuracy.

To address this issue, we introduce \textbf{MathSight}, a new university-level multimodal mathematical reasoning benchmark designed to disentangle and quantify the influence of visual information. Each problem instance is presented under multiple visual variants, such as the original figure, a hand-drawn version, and a photo-captured version, enabling a controlled and systematic analysis of how variations in visual representation impact reasoning. In addition, we incorporate a text-only condition without any image input to further evaluate whether VLMs can solve problems purely through linguistic reasoning, independent of visual cues. Moreover, we also construct a university-level text-only benchmark to assess the reasoning capability of large language models (LLMs) under comparable difficulty settings, serving as a complementary reference for the multimodal analysis.


Through extensive experiments on multiple state-of-the-art VLMs (e.g., Gemini-2.5-pro, GPT-5, Claude-Opus, Claude-Sonnet, and Qwen3-vl), we uncover a revealing trend: the contribution of visual information to mathematical reasoning diminishes as problem difficulty increases. In our university-level MathSight benchmark, the three visual variants—original, hand-drawn, and photo-captured variants—yield no statistically significant differences in model performance (See Figure~\ref{fig:motivation_example}). Even more intriguingly, Qwen3-VL achieves 50.53\% accuracy without any image input, surpassing not only its own multimodal variants (\textit{original}: 40.85, \textit{hand-drawn}: 38.73, \textit{photo-captured}: 39.33) but also outperforming GPT-5 (45.39) under identical conditions.
This striking result indicates that high-level reasoning in current VLMs may rely more on textual priors and linguistic abstraction than on genuinely grounded visual understanding. 

The main contributions of this paper are summarized as follows:

\begin{itemize}
   \item \textbf{A new benchmark for vision-grounded mathematical reasoning.} We introduce \textbf{MathSight}, a university-level multimodal mathematical reasoning benchmark specifically designed to disentangle and quantify the influence of visual information. It enables systematic evaluation of VLMs under multiple controlled visual conditions.

    \item \textbf{A comprehensive analysis of visual sensitivity in VLMs.} By evaluating several state-of-the-art models (e.g., Gemini 2.5 Pro, GPT-5, Claude-Opus, Claude-Sonnet, and Qwen3-VL) across original, hand-drawn, photo-captured, and text-only settings, we reveal that the impact of visual input decreases as the problem difficulty increasing.

    \item \textbf{Empirical insights into the limits of current multimodal reasoning.} Our findings show that models often achieve comparable or even superior performance without image input—highlighting a fundamental gap between true visual understanding and language-driven reasoning in current VLMs.
   
\end{itemize}

\begin{table*}[t!]
\centering
\caption{Existing Benchmarks, where "Visual.Variant" indicates that visual input has different variants, "Grad. Level" presents graduate-level problems and "Pro.Q" stands for proving questions, i.e., questions that cannot be simply verified against a standard answer.}
\label{tab:bench_list}
\begin{adjustbox}{max width=0.99\linewidth}  
\begin{tabular}{llccccc}
\specialrule{.16em}{0pt}{.65ex}
\multirow{2}{*}{Benchmark} & \multirow{2}{*}{Grad. Level} &\multirow{2}{*}{Pro.Q} &\multicolumn{3}{c}{Visual.Variant} & \multirow{2}{*}{Image.Size} \\ \cline{4-6}
& &  & Original & Hand-drawn & Photo-captured & \\
\specialrule{.05em}{.4ex}{.65ex}
TheoremQA \cite{chen2023theoremqa} & \xmark & \xmark &\cmark &\xmark &\xmark &\xmark\\ 

MathVista \cite{lu2023mathvista} & \xmark &\xmark &\cmark &\xmark &\xmark &\xmark\\ 
Scibench \cite{Wang:23SciBench} & \xmark & \xmark &\cmark &\xmark &\xmark &\xmark\\ 
QRData \cite{liu2024qrdata} &\xmark &\xmark &\cmark &\xmark &\xmark &\xmark\\ 
MATH-Vision \cite{wang2024math-vision}  & \xmark &\xmark &\cmark &\xmark & \xmark &\xmark\\ 
$MMMU_{math}$\cite{Yue:24MMMU}  &\xmark &\xmark &\cmark &\xmark &\xmark&\xmark\\ 
U-Math \cite{ko2024umath} & \xmark &\xmark &\cmark &\xmark &\xmark &\xmark\\ 
Dynamath \cite{Zou:24DynaMath} & \xmark &\xmark &\cmark &\xmark &\xmark &\xmark\\ 
MathCheck \cite{zhou2024your} & \xmark &\xmark &\cmark &\xmark &\xmark &\xmark\\ 
MathVerse \cite{zhang2025mathverse} & \xmark &\xmark &\cmark &\xmark &\xmark &\xmark\\ 
MathFlow \cite{chen2025mathflow} & \xmark &\xmark &\cmark &\xmark &\xmark &\xmark\\ 
\specialrule{.05em}{.4ex}{.65ex}
\textbf{MathSight} (our work)  &\cmark &\cmark &\cmark &\cmark &\cmark &\cmark\\
\specialrule{.16em}{0pt} {.65ex}
\end{tabular}
\end{adjustbox}
\end{table*}

\section{MathSight} 


In this section, we introduce \textbf{MathSight}, a benchmark designed to investigate how visual information influences mathematical reasoning in vision-language models (VLMs). By varying the visual input—through original, hand-drawn, and photo-captured figures—while keeping the problems fixed, MathSight enables controlled analysis of visual contributions to reasoning performance and provides new insights into multimodal understanding. In the following, we describe the dataset construction, dataset composition, data analysis, and metrics for the logical consistency of MathSight in detail.

\subsection{Dataset Construction}
The pipeline of dataset construction includes three main phases: document selection, QA pair generation, and image augmentation.

\subsubsection{Document Selection.} 
The dataset was collected from proprietary, non-public PDF documents provided by a collaborating company. These documents have never been made publicly available, ensuring that no evaluated model has been exposed to the data during pretraining.

We performed multiple rounds of manual screening. First, non-mathematical documents and those below the university level were excluded. Then, domain knowledge was utilized to classify and curate high-quality problem sets. Only problems containing a complete \texttt{question}, \texttt{answer/solution}, and \texttt{groundtruth} were retained.

\subsubsection{QA Pair Generation.}
After filtering the PDFs, we used \texttt{GPT-4o} to extract problems and convert them into our standardized \texttt{JSONL} format. A reusable extraction strategy was designed to guide the model, which can consistently produce well-formatted outputs. 

The strategy consists of two parts: 1) a standard formatting guide specifying required fields (\texttt{question}, \texttt{answer/solution}, \texttt{groundtruth}, and optional \texttt{image\_path}), and 2) error-handling rules such as skipping incomplete items and logging issues. In most cases, each multimodal problem is associated with a unique image. A few exceptions exist where multiple questions share a single visual reference.

Following extraction, we used \texttt{GPT-4o} again to flag questions below the university level. These were re-checked by human experts. Only problems verified to meet undergraduate-level or graduate-level difficulty were retained. 

Finally we manually screened over 20,000 source files to extract approximately 2,600 math problems, then selected 661 high-quality, multimodal math reasoning items and 1,387 text-only problems based on clarity, solvability, and modality diversity. While the final benchmark is relatively small, it is intentionally quality-focused and covers diverse categories to ensure reliable evaluation.

\subsubsection{Image Augmentation.}
Last but not least, we conduct augmentation for the images in multimodal problems manually. In order to simulate a real situation, we draw all the images with 5 graduate students whose handwriting is different from one another. Then the hand-drawn figures are taken as photos to use. To imitate photos that people take from paper materials, we print all the images and collect the photo-captured version of the image. Then the two variants of the original image possess large sizes for use with a mobile phone camera. These raw photos of variants are cropped and reduced in size to keep consistent with the original images.

\subsection{Data Analysis}


This section exposes the data analysis of MathSight from different perspectives.

\subsubsection{Multimodal Coverage.}
Our dataset contains 661 university-level multimodal problems, 
among which 603 are at the graduate-level, 
This number at the graduate level significantly exceeds that of other benchmarks. The right part of Figure~\ref{fig:motivation_example} presents a representative multimodal example in MathSight, which is a geometric problem-solving task involving hypothesis reasoning and spatial visualization. For such problems, a model is considered correct only if it provides the exact final answer, accompanied by a logically consistent reasoning process that correctly understands the given graph. Partial answers, incorrect intermediate steps, or misinterpretations of the visual information may lead to wrong answers.


\subsubsection{Subject Distribution.}
We construct university-level problems across six core mathematical domains: Calculus, Algebra, Analysis, Probability \& Statistics, Discrete Mathematics, and Applied Math. These domains encompass both foundational undergraduate courses and more advanced or specialized areas, such as mathematical logic and topology, which are essential for assessing deep mathematical reasoning capabilities. 

Calculus and Applied Math constitute the largest portions of the dataset, accounting for 82.15\% and 11.8\%. The remaining domains contribute meaningfully to the coverage. 
The distribution of undergraduate-level and graduate-level problems within each subject allows us to analyze the models' capabilities across varying levels. 
These distributions would help provide valuable insight into the areas where current large models excel or struggle.

\subsubsection{Proving Problems.}
Our dataset contains a total of 30 proving problems, accounting for 4.54\% of all multimodal questions. Among them, 29 proving problems are at the graduate level, while only one is at the undergraduate level. We intentionally searched proving problems for MathSight because existing benchmarks exhibit deficiencies on such tasks. The performances of large models on these tasks are worth exploring. Moreover, proving problems are distributed across all six domains, enabling us to evaluate models’ reasoning abilities from multiple disciplinary perspectives. 


\subsection{Metrics for Logical Consistency}
MathSight dataset includes a considerable proportion of proving problems. To evaluate the logical reasoning of large models on these problems, we propose several metrics inspired by recent studies \cite{fu2025deepthinkconfidence, kang2025scalable}. 


\paragraph{Token Confidence.}
Following \cite{fu2025deepthinkconfidence}, the token confidence $C_i$ at position $i$ is defined as
\begin{equation}
C_i \;=\; -\frac{1}{k} \sum_{j=1}^{k} \log P_i(j),
\label{eq:token_confidence}
\end{equation}
where $P_i(j)$ denotes the predicted probability of the $j$-th token in the top-$k$ candidates at step $i$,
and $k$ is the number of top tokens considered. High $C_i$ corresponds to a more peaked distribution
and thus greater model certainty.

\paragraph{Sliding-window Group Metrics.}
Given token confidences $\{C_i\}_{i=1}^N$ (Eq.~\ref{eq:token_confidence}),  
where $N$ is the total number of generated tokens in the model’s answer/solution,  
we form $M$ overlapping windows 
$\mathcal{I}_j=[a_j,b_j]\subseteq\{1,\ldots,N\}$ using window size $W$ and step $S$
(with tail mode ``full''). For each window, we compute the local group mean:
\begin{equation}
g_j \;=\; \frac{1}{|\mathcal{I}_j|}\sum_{i\in\mathcal{I}_j} C_i.
\end{equation}

We report three summary metrics over $\{g_j\}_{j=1}^M$:

\paragraph{Group Overall Mean(GOM).}
We compute the average of all group means $g_j$ to reflect the overall local confidence level 
across the entire generated answer/solution:
\begin{equation}
\bar{g} \;=\; \frac{1}{M}\sum_{j=1}^{M} g_j.
\label{eq:overall_mean}
\end{equation}


\paragraph{Group Standard Deviation(GSD).}
We compute the (population) standard deviation of all group means to measure the fluctuation 
of confidence along the trace:
\begin{equation}
\sigma_g \;=\; \sqrt{\frac{1}{M}\sum_{j=1}^{M}\bigl(g_j-\bar{g}\bigr)^2}.
\label{eq:std}
\end{equation}

\paragraph{Group Coefficient of Variation(GCV).}
We compute the coefficient of variation  to assess the relative fluctuation of 
confidence independent of its absolute level:
\begin{equation}
\mathrm{CV}_g \;=\; \frac{\sigma_g}{\bar{g}}\;.
\label{eq:cv}
\end{equation}

\begin{figure*}[t]
\centering
\includegraphics[width=0.32\textwidth]{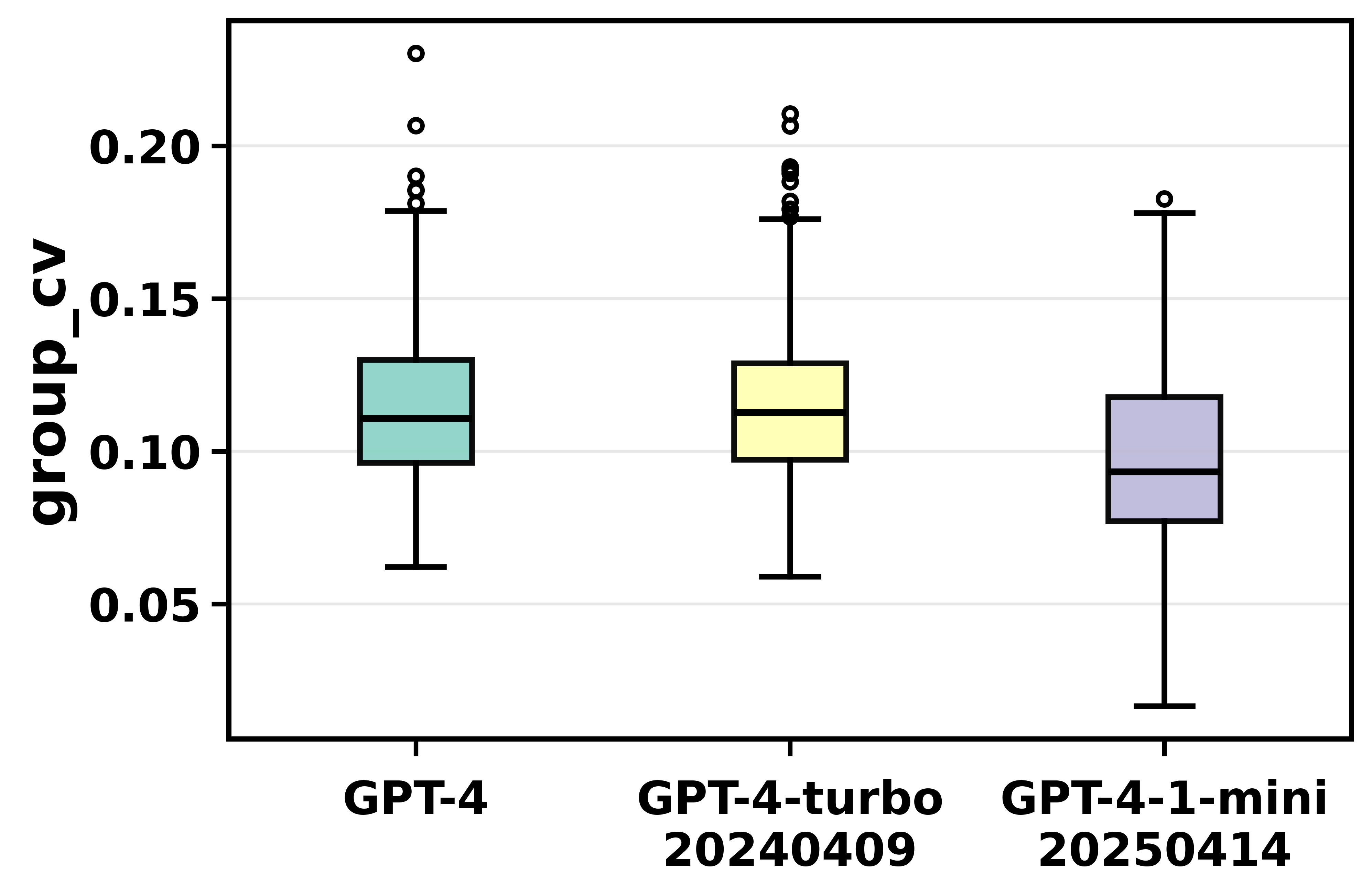}\hfill
\includegraphics[width=0.32\textwidth]{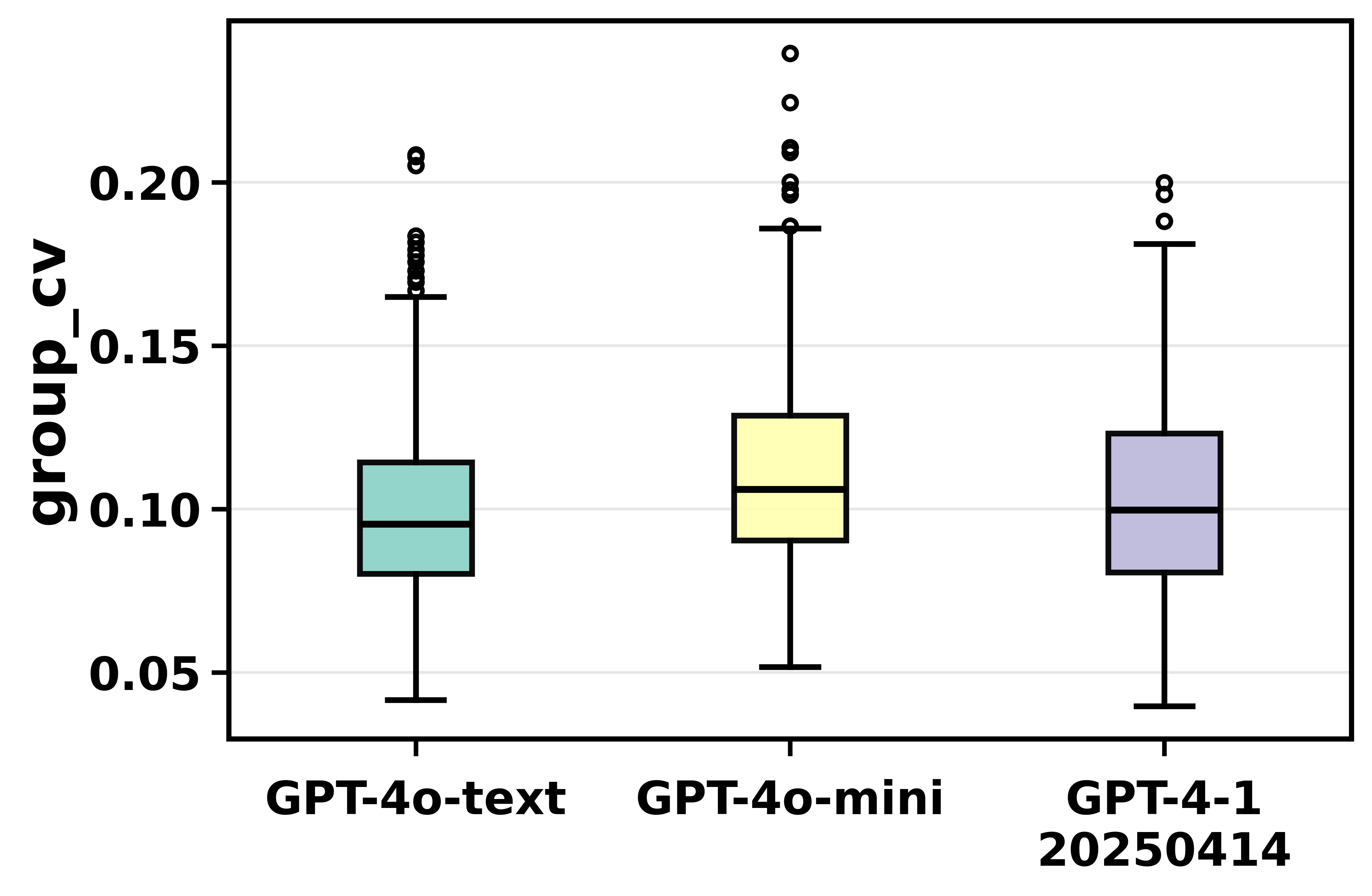}\hfill
\includegraphics[width=0.32\textwidth]{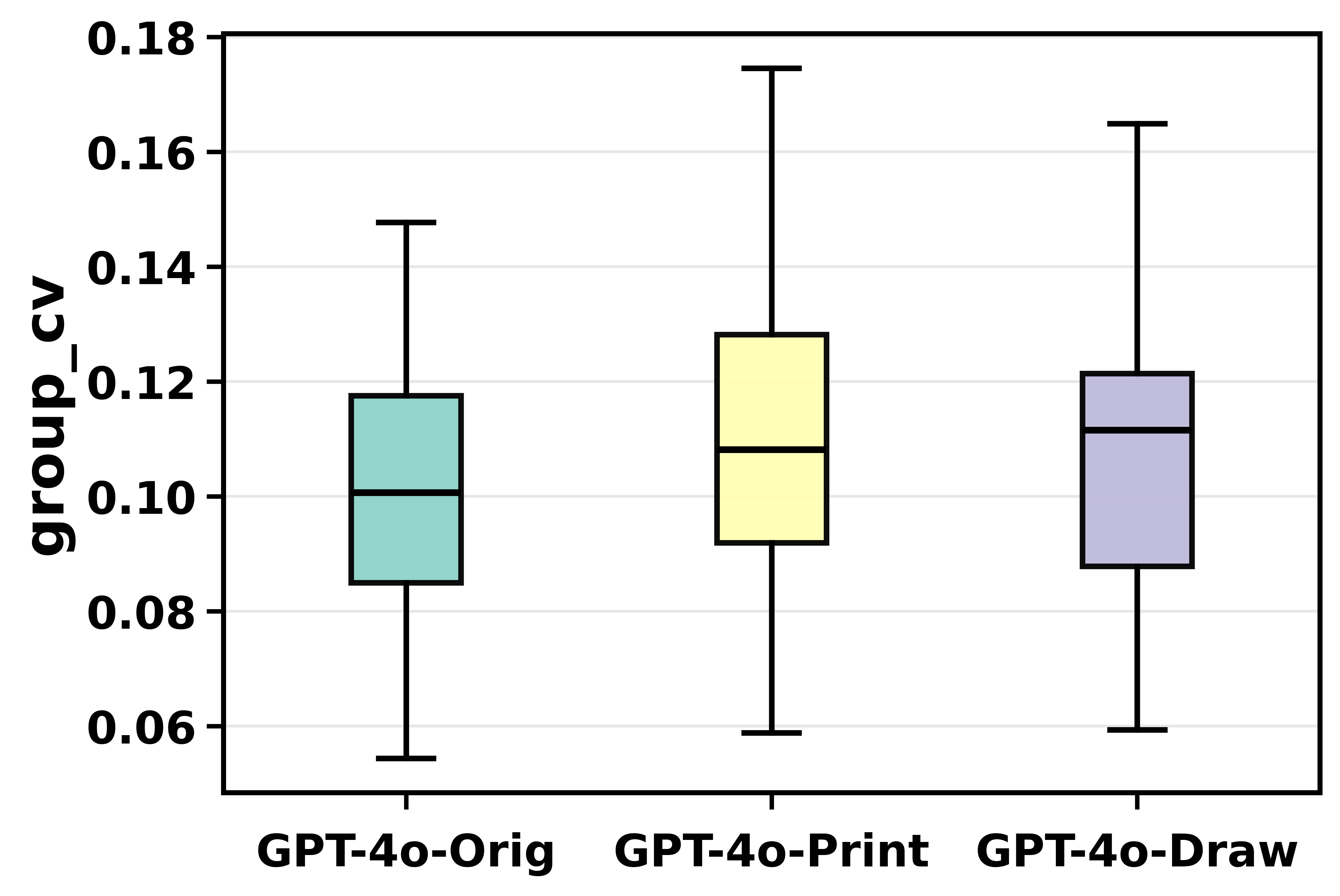}
\caption{Box plots of group coefficient of variation (group\_CV) across different models and $k$ settings. 
Each box covers the 95\% central region of group\_CV values, the horizontal line inside the box marks 
the median group\_CV, and the points outside indicate a small number of outliers.
Left: text-only model with $k=5$. Middle: text-only model with $k=20$. 
Right: multimodal GPT-4o variants with $k=20$.}
\label{fig:group_cv_boxplots}
\end{figure*}




\section{Evaluation and Results} 
In this section, we report a comprehensive evaluation of various Vision-Language Models (VLMs) on the proposed \textit{MathSight} benchmark.
Our experiments aim to examine how visual factors influence mathematical reasoning at the university level, with particular attention to \textit{image variation} (e.g., type and size) and \textit{problem diversity} across different mathematical categories.
We systematically analyze model performance across multiple visual conditions and reasoning domains to uncover the extent to which current models truly leverage visual information in complex mathematical problem solving.

\subsection{Evaluated Models}

We evaluate a diverse set of state-of-the-art Vision-Language Models (VLMs) that represent the current frontier of multimodal reasoning from open-source and closed-source. Our evaluation includes Gemini-2.5-Pro~\cite{Comanici:25Gemini2.5}, GPT-5, the Claude family (\textit{Claude-3.7-Sonnet}, \textit{Claude-4-Sonnet}, and \textit{Claude-Opus})~\cite{anthropic2025claude37,anthropic2025claude4}, Qwen2.5-VL (7B, 3B)~\cite{bai2025qwen25vl}, Doubao-1.5-Vision-Pro~\cite{guo2025seed15vl}, GLM-4.1V-Thinking~\cite{Hong:25GLM41VThinking}, InternVL3 (78B)~\cite{Zhu:25InternVL3}, and Qwen3-VL-235B-A22B.

To further disentangle the contribution of visual information, we also evaluate the text-only variant Qwen3-235B-A22B, which shares the same underlying language backbone as Qwen3-VL but operates purely on textual input. 
This comparison allows us to quantify how much of the reasoning capability stems from linguistic priors versus vision-grounded understanding. 
All models are tested in a zero-shot setting using their publicly available APIs or official implementations, ensuring reproducibility and fair comparison.

\subsection{Experimental Setup}
To ensure the reproducibility of our experiments, we describe the evaluation metric and the prompt templates used during model inference and evaluation in detail.

We report \textbf{accuracy (ACC)} for non-proof questions. 
For proof problems, we evaluate \textbf{logical consistency} of the reasoning process instead of matching a specific expression. 
Specifically, we use three complementary metrics---\textbf{GOM}, \textbf{GSD}, and \textbf{GCV}---to capture  overlap with essential logical dependencies,  step-level deviation from sound inference, and stability of consistency across the proof.

All experiments are conducted in a zero-shot setting. We obtain model responses from two sources: 1) API-based access to proprietary or cloud-hosted models, and 2) local inference using open-source models with the vLLM\cite{vllm} library. 
Final accuracy is computed by comparing model outputs with ground truth using  mathematical equivalence, depending on the task type.
We will release our benchmark on GitHub in the near future.



\subsection{Overall Results}

Table~\ref{tab:mm_res} presents the overall performance of representative vision-language models (VLMs) on the MathSight benchmark under three visual conditions: \textit{original}, \textit{hand-drawn}, and \textit{photo-captured} image variants. Across both closed-source and open-source model, we observe a consistent trend: although accuracy slightly varies across image variants, the difference remain marginal, suggesting that visual appearance has only a limited impact on university-level mathematical reasoning. This finding indicates that current VLMs may tend to rely predominantly on textual cues or memorized symbolic patterns rather than genuinely integrating visual understanding into the reasoning process. 

Interestingly, several models exhibit slightly higher accuracy on the \textit{hand-drawn} and \textit{photo-captured} variants, indicating that imperfect or naturalistic visuals do not necessarily impede comprehension. Instead, such inputs may encourage models to depend more on symbolic and relational reasoning rather than superficial visual pattern matching. This finding further supports the hypothesis that contemporary VLMs predominantly rely on textual abstraction rather than genuinely grounded visual understanding. 

Moreover, MathSight, a university-level benchmark for assessing advanced multimodal mathematical reasoning, remains highly challenging. Even the state-of-the-art model, GPT-5, achieves only 45.39\% accuracy, indicating a considerable gap from human-level competence. Although closed-source models such as Gemini and Claude generally outperform open-source counterparts, the performance gap is notably small.
In fact, we find that tasks correctly solved by one model are often solved by most others, whereas those failed by one tend to be universally difficult across models. This pattern suggests that current VLMs occupy a narrow reasoning space and exhibit highly correlated failure modes, revealing a lack of complementary multimodal capabilities across architectures.

\begin{table}[t!]
\centering
\caption{Overall evaluation results of Vision-Language Models (VLMs) on the \textbf{MathSight} benchmark. }
\label{tab:mm_res}
\renewcommand{\arraystretch}{1.15}
\begin{adjustbox}{max width=0.99\linewidth}
\begin{tabular}{llcc}
\specialrule{.16em}{0pt} {.65ex}
Model & V.Orig & V.Draw & V.Photo \\ 
\specialrule{.05em}{.4ex}{.65ex}
\multicolumn{4}{c}{\textit{\textbf{Closed-source Models}}}\\
\hdashline
\midrule
Claude-3.7-sonnet-20250219 & 36.80 & 39.64 & 40.39 \\
Claude-3.7-sonnet-20250219-thinking & 34.80 & 37.52 & 41.15 \\
Claude-sonnet-4-20250514 & 36.16 & 37.67 & 40.70 \\ 
Claude-sonnet-4-20250514-thinking & 36.61 & \textbf{41.60} & \textbf{41.45} \\ 
Claude-opus-4-20250514 & 36.61 & 37.97 & 39.79 \\ 
Claude-opus-4-20250514-thinking & 35.55 & 41.45 & 42.36 \\ 
Gemini-2.5-pro-preview-06-05 & 37.07 & 38.73 & 40.85 \\ 
GPT-4o(2024/05/13) & 36.91 & 36.46 & 37.52 \\ 
GPT-5 & \textbf{45.39} & \underline{41.50} & 40.85 \\ 
\midrule
\multicolumn{4}{c}{\textit{\textbf{Open-source Models}}}\\
\hdashline
\midrule
InternVL3-78B & 39.64 & 39.49 & \underline{41.30} \\ 
GLM-4.1V-9B-Thinking & 35.85 & 37.37 & 38.58 \\ 
Qwen2.5-vl-7B & 34.64  & 34.95 & 36.31 \\ 
Qwen2.5-vl-3B & 33.74 & 36.61 & 36.61 \\ 
Qwen3-VL-235B-A22B & \underline{40.85} & 38.73 & 39.33 \\ 
\specialrule{.16em}{0pt} {.65ex}
\end{tabular}
\end{adjustbox}
\end{table}

\begin{table*}[t]
\centering
\caption{Evaluation results of Vision-Language Models (VLMs) under different subjects. (Note that subject abbreviations indicate: Applied. = Applied Math, Pro.\&Sta. = Probability \& Statistics, Discrete. = Discrete Mathematics). The \textbf{bold} and \underline{underline} numbers represent the best and the second best results, respectively. 
 }
\label{tab:category_res}
\begin{adjustbox}{max width=0.8\linewidth}
\begin{tabular}{llcccccc}
\specialrule{.16em}{0pt} {.65ex}
\multirow{2}{*}{Model} & All & Calculus & Algebra & Applied. & Pro.\&Sta. & Analysis & Discrete.  \\
&661 & 543 & 7 & 78 & 21 & 6 & 6 \\ 
\specialrule{.05em}{.4ex}{.65ex}
\multicolumn{8}{c}{\textit{\textbf{Closed-source Models}}}\\
\hdashline
\midrule
Claude-3.7-sonnet-20250219 & 36.76 & 29.10 & 71.43 & 73.08 & 76.19 & 50.00 & 66.67 \\
Claude-3.7-sonnet-20250219-thinking & 34.80 & 27.07 & 71.43 & 69.23 & 80.95 & 50.00 &66.67 \\
Claude-sonnet-4-20250514 & 36.16 & 28.55 & 71.43 & 71.79 & 80.95 & 33.33 &66.67 \\ 
Claude-sonnet-4-20250514-thinking & 37.97 & 30.57 & 71.43 & 70.51 & \underline{85.71} & 50.00 &66.67 \\ 
Claude-opus-4-20250514 & 37.52 & 29.65 & 71.43 & \underline{74.36} & \underline{85.71} & 33.33 & 66.67 \\ 
Claude-opus-4-20250514-thinking & 35.55 & 28.36 & 71.43 & 69.23 & 71.43 & 50.00 &66.67 \\ 
Gemini-2.5-pro-preview-06-05 & 37.07 & 28.73 & \underline{85.71} & 73.08 & \textbf{90.48} & 50.00 &66.67  \\ 
GPT-4o(2024/05/13) & \underline{38.43} & \underline{31.12} & \underline{85.71} & 69.23 & 80.95 & \underline{66.67} & 66.67 \\ 
GPT-5 & \textbf{45.39} &37.94 &85.71 &79.49 &85.71 &66.67 &66.67\\
\midrule
\multicolumn{8}{c}{\textit{\textbf{Open-source Models}}}\\
\hdashline
\midrule
InternVL3-78B & \textbf{39.64} & \textbf{32.78} & \textbf{71.43} & \textbf{70.51} & \underline{76.19} & \textbf{66.67} &66.67 \\ 
GLM-4.1V-9B-Thinking & \underline{35.85} & \underline{28.36} & \underline{71.43} & \underline{70.51} & \underline{76.19}  & \underline{50.00} &66.67 \\ 
Qwen2.5-vl-7B & 34.64 & 27.44 & \underline{71.43} & 66.67 & \textbf{80.95} & 33.33 &66.67 \\ 
Qwen2.5-vl-3B & 33.74 & 25.97 & \underline{71.43} & 69.23 & \underline{76.19}  & \underline{50.00} &66.67  \\ 
Qwen3-VL-235B-A22B & \underline{40.85} &33.70 &85.71 &71.79 &85.71 &50.00 &66.67 \\
\specialrule{.16em}{0pt} {.65ex}
\end{tabular}
\end{adjustbox}
\end{table*}

\subsection{Results on Various Categories}

To further analyze how visual information interacts with different forms of reasoning, we evaluate VLMs' performance on the \textit{original-variant} setting across various mathematical categories, including Calculus, Algebra, Applied Mathematics, Probability \& Statistics, Analysis, and Discrete Mathematics. 
As shown in Table~\ref{tab:category_res}, we observe substantial variation across categories, revealing that visual dependence is not uniform but highly task-specific. 
Models generally achieve the highest accuracy on \textit{algebraic} and \textit{probabilistic} problems, where reasoning can often proceed from symbolic or textual cues alone. 
In contrast, performance on \textit{calculus} and \textit{analysis} mathematics is consistently lower, indicating that tasks requiring conceptual abstraction or geometric interpretation remain challenging for current VLMs.
Interestingly, \textit{applied mathematics} and \textit{probability} problems—where visual representations are closely aligned with textual descriptions—exhibit comparatively higher accuracy, suggesting that models can partially exploit structured visual layouts when visual–textual alignment is strong. 
However, this advantage likely stems from surface-level correlation matching rather than genuine visual reasoning, underscoring that current VLMs rely heavily on linguistic priors rather than grounded visual understanding.



\subsection{Results on Different-size Images}

\begin{figure*}[t!]
    \centering
    \includegraphics[width=0.9\linewidth]{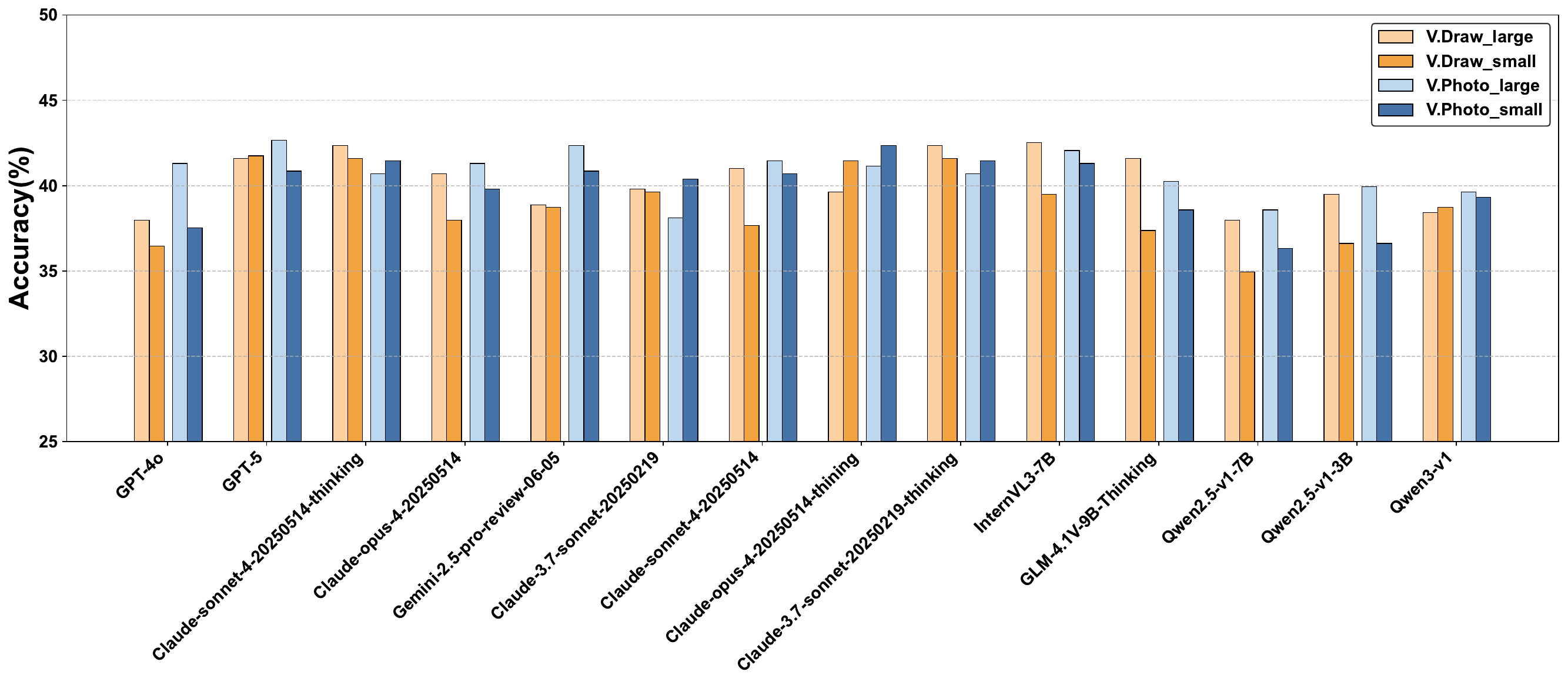}
    \caption{Evaluation results of Vision-Language Models (VLMs) on images of different sizes.}
    \label{fig:image_size_res}
\end{figure*}

To investigate the sensitivity of Vision-Language Models (VLMs) to visual scale and resolution, we further evaluate their performance under four controlled visual configurations: \textit{large} and \textit{small} versions of both \textit{hand-drawn} and \textit{photo-captured} figures. The results are presented in Figure~\ref{fig:image_size_res}. Across all models, we observe that \textbf{variations in image size have minimal influence on overall accuracy}. For most VLMs, the difference between large and small images within the same visual type remains within a narrow range of 1–2 percentage points, far smaller than the variance typically observed across problem categories or modalities. This stability suggests that current VLMs rely little on fine-grained visual detail when solving mathematical problems. Instead, their reasoning appears to be dominated by textual understanding and symbolic pattern recognition extracted from the accompanying problem statement.

Interestingly, for some models such as \textbf{Gemini 2.5 Pro} and \textbf{Claude 3.7 Sonnet}, performance with smaller hand-drawn or photo images is even slightly higher than with larger ones. This counterintuitive behavior indicates that resizing visual inputs does not degrade, and may occasionally regularize, model predictions—possibly because smaller images implicitly downweight the visual embedding relative to the text, thereby reducing visual noise. 

Taken together, these results suggest that current VLMs display a high degree of \textbf{scale invariance} yet a low degree of \textbf{visual reliance}. In other words, altering the scale of the visual input scarcely impacts reasoning performance, implying that these models engage only superficially with visual information and instead depend predominantly on linguistic priors.


\subsection{Text-Only vs. Multimodal Reasoning on VLMs}

To directly assess the contribution of visual input, we evaluate \textbf{Qwen3-VL-235B-A22B} under two settings: a \textit{text-only} setting and a \textit{multimodal} setting. In the text-only setting, Qwen3-VL only receives the full problem statement but no accompanying visual content, allowing us to isolate whether the visual modality provides complementary grounding or merely redundant information. 
In the multimodal configuration, we further examine three types of visual inputs such as the \textit{original}, \textit{hand-drawn}, and \textit{photo-captured} variants to analyze how different visual representations influence the performance of university-level mathematical reasoning.

\begin{figure}
    \centering
    \includegraphics[width=0.9\linewidth]{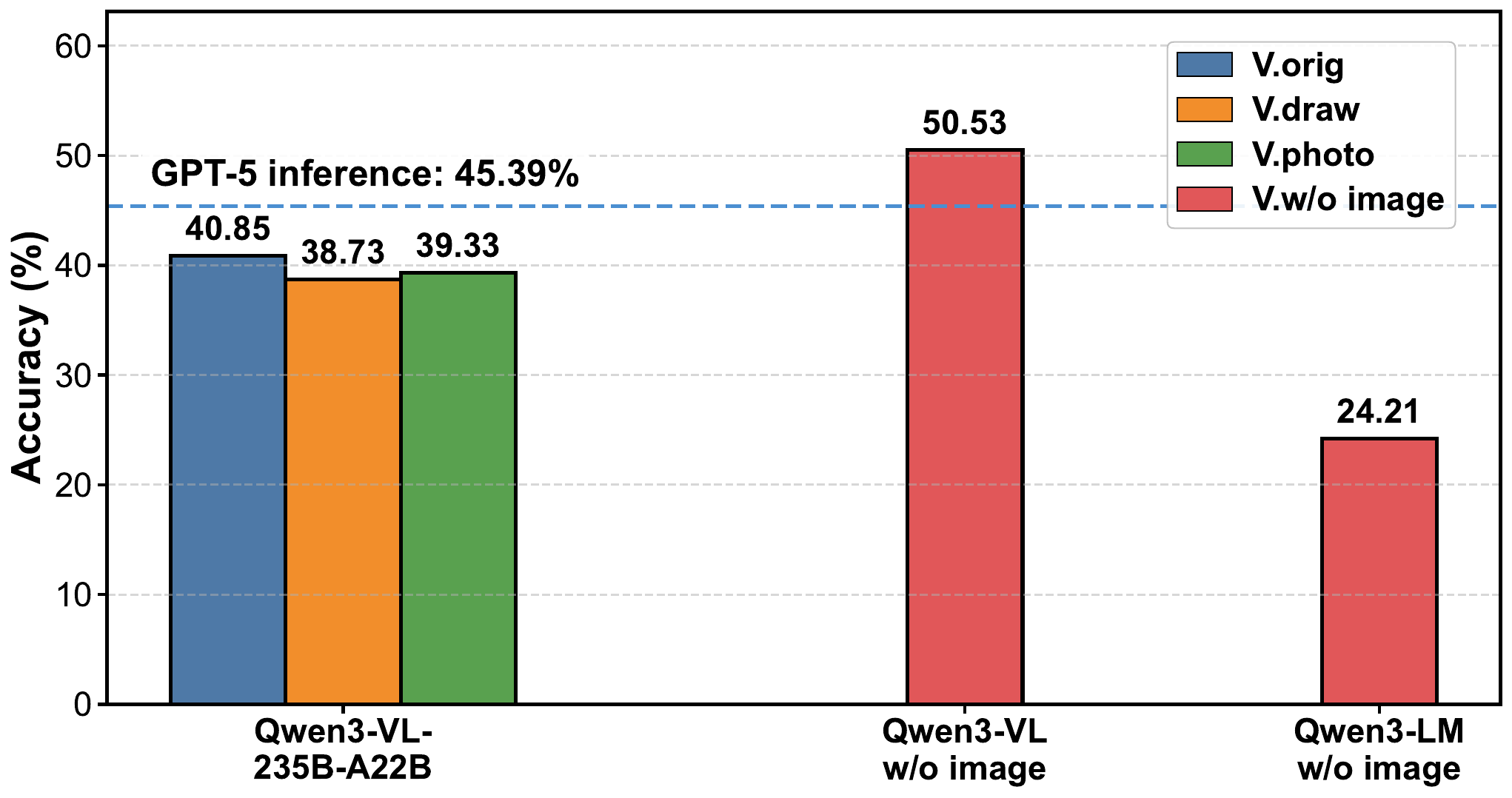}
    \caption{The result comparison between text-only and different multimodal inputs.}
    \label{fig:text_only_with_images}
\end{figure}

As shown in Figure~\ref{fig:text_only_with_images}, we observe that \textbf{removing the image input leads to a substantial performance improvement}, boosting accuracy from 40.85\% to 50.53\% and even outperforming the state-of-the-art GPT-5 (45.39\%). 
This counterintuitive result suggests that current VLMs may not truly engage in visual reasoning but rather treat visual inputs as noisy or even misleading signals when solving mathematically complex problems. 
The improvement in the text-only setting indicates that current VLMs may rely heavily on linguistic priors and symbolic patterns learned during pretraining, rather than deriving meaningful information from visual input.
This finding highlights a fundamental limitation of current VLMs' architectures: the presence of an image does not guarantee vision-grounded understanding, especially in university-level mathematics.

In addition, we extend the text-only evaluation to the \textbf{Qwen3-LM} model, which shares the same language backbone as Qwen3-VL but lacks any visual processing capability. Interestingly, its accuracy drops significantly to \textbf{24.21\%}, far below both the multimodal and text-only variants of Qwen3-VL. This sharp contrast indicates that the superior text-only performance of Qwen3-VL does not stem merely from its linguistic backbone but from its training as a multimodal model, which appears to internalize certain structural priors that generalize even without visual input.

\subsection{Further Analysis}

To further investigate how visual variations influence reasoning stability, we analyze the consistency of model predictions across the three image variants (\textit{original}, \textit{hand-drawn}, and \textit{photo-captured}). Table~\ref{tab:image_variants_consistency} reports the distribution of answer combinations for GPT-5 and Qwen3-VL-235B-A22B.

We observe that the performance consistency across different image variants is extremely high for both GPT-5 and Qwen3-VL. 
More than \textbf{80\%} of all cases remain fully consistent, either all correct (\textit{A\cmark B\cmark C\cmark}) or all incorrect (\textit{A\xmark B\xmark C\xmark}), while only about \textbf{18\%} show any variation across visual versions. 
Such strong consistency implies that model predictions are largely insensitive to changes in visual form, suggesting that performance is driven primarily by linguistic or symbolic reasoning rather than genuine visual grounding. In other words, whether the accompanying image is original, hand-drawn, or captured in a photo has little influence on the outcome: the models effectively ``read'' but do not ``see''.
This finding underscores that current VLMs treat visual input as an auxiliary cue rather than an integral reasoning component, revealing a fundamental limitation in their capacity for truly vision-conditioned mathematical understanding.

To further clarify why visual input does not consistently improve VLM reasoning, we conducted additional controlled comparisons across Qwen3, Qwen2.5, and multiple LLM baselines.
Across all architectures, a consistent performance ordering emerges:
{\small
\[
\begin{aligned}
\text{Qwen-VL (w/o image)} \;>\; \text{Qwen-VL (text + image)} \\
\;>\; \text{Qwen-LM (caption)} \;>\; \text{Qwen-LM (text-only)}.
\end{aligned}
\]
}
This monotonic pattern holds for both the Qwen3 and Qwen2.5 families, as shown in Table~\ref{tab:qwen-series without caption}.
These results lead to three key observations:
\begin{enumerate}
    \item \textbf{Visual input often reduces accuracy.}
    For both Qwen3-VL and Qwen2.5-VL, adding the image consistently lowers accuracy compared to the text-only setting.
    This indicates that current visual encoders may introduce irrelevant or noisy perceptual tokens, which interfere with the linguistic reasoning pathway.

    \item \textbf{VLMs lack effective modality selection.}
    Models fail to determine when visual information is useful and when it should be ignored.
    When the image is unnecessary or only weakly informative, the model still processes visual tokens, leading to degraded performance.

    \item \textbf{LLM + captions cannot replace true multimodal modeling.}
    Even with image captions, LLMs remain substantially weaker than VLMs.
    This confirms that the gap does not stem from missing textual cues, but rather from modality alignment challenges intrinsic to current VLM architectures.
\end{enumerate}




\begin{table}[t!]
\centering
\caption{The percentage of different answer combinations on three image versions, GPT-5 and Qwen3-VL-235B-A22B as examples. Note that A denotes \textit{original variant}, B represents \textit{photo-captured variant}, and C corresponds to \textit{hand-drawn variant}.}
\label{tab:image_variants_consistency}
\renewcommand{\arraystretch}{1.15}
\begin{adjustbox}{max width=0.99\linewidth}
\begin{tabular}{llc}
\specialrule{.16em}{0pt} {.65ex}
Answer Comparison & GPT-5 & Qwen3-VL-235B-A22B \\ 
\specialrule{.05em}{.4ex}{.65ex}
A\cmark B\cmark C\cmark & 34.19\% &30.56\%\\ 
A\cmark B\cmark C\xmark & 3.48\% &3.48\%  \\ 
A\cmark B\xmark C\cmark & 2.87\% &2.87\% \\ 
A\cmark B\xmark C\xmark & 4.84\% &3.93\% \\ 
A\xmark B\cmark C\cmark & 1.36\% &2.42\%  \\ 
A\xmark B\cmark C\xmark & 2.72\% &2.87\%  \\ 
A\xmark B\xmark C\cmark & 2.42\% &2.87\%  \\ 
A\xmark B\xmark C\xmark & 48.11\% &50.98\%\\ 
\specialrule{.16em}{0pt} {.65ex}
\end{tabular}
\end{adjustbox}
\end{table}



\begin{table}[t]
\centering
\small
\caption{Accuracy (\%) of Qwen3-235B-A22B (Qwen3-VL-235B-A22B / Qwen3-LM-235B-A22B) and Qwen2.5-3B (Qwen2.5-VL-3B / Qwen2.5-LM-3B) under different input settings.}
\label{tab:qwen-series without caption}
\begin{tabular}{lcc}
\toprule
\textbf{Input setting} & \textbf{Qwen3} & \textbf{Qwen2.5} \\
\midrule
VL, text-only (no image)      & 50.53 & 61.72 \\
VL, with image                & 40.85 & 33.74 \\
LM, with image caption        & 28.74 & 25.42 \\
LM, text-only (no caption)    & 24.05 & 24.66 \\
\bottomrule
\end{tabular}
\end{table}

\section{Related Work}

In order to study the effects of mathematical diagrams on problem-solving, various multimodal benchmarks are proposed \cite{chen2023theoremqa, lu2023mathvista, Gupta:24Polymath, Wang:23SciBench, zhang2024geoeval, wang2024math-vision, Yue:24MMMU, liu2024qrdata, Zou:24DynaMath, zhou2024your, ko2024umath}. Recent multimodal mathematical benchmarks are not only focus on geometric figures \cite{zhang2024geoeval} but also broaden to wider range, such as venn diagrams, spatially-related layouts \cite{Gupta:24Polymath}. MMMU \cite{Yue:24MMMU} offers multi-discipline reasoning, and its limited math subset consists of graph functions and mathematical notation images. U-MATH \cite{ko2024umath} is composed of 1100 enclosed open-ended university-level problems and an extra dataset to testify the judging abilities of LLMs. Similar to UGMathBench \cite{xu2025ugmathbench}, DYNAMATH \cite{Zou:24DynaMath} is also a dynamic benchmark but with visual information, which includes 501 high-quality, multi-topic seed questions. Though these benchmarks cover multimodal input, they lack detailed exploration on the effectiveness of the image attributes. In MathSight, we not only collected diverse-category problems but also augmented the image into different variants, which is an outstanding superiority for visual understanding research.

\section{Conclusion}
We introduced MathSight, a university-level benchmark designed to isolate and measure the value of visual information in multimodal mathematical reasoning. Each problem is paired with multiple visual variants (original, hand-drawn, photo) and a text-only condition, enabling controlled comparisons. We also provide a difficulty-matched, text-only benchmark to calibrate language-model reasoning. Across state-of-the-art VLMs, our results are clear: as problems get harder, the benefit of images shrinks. Performance differences among visual variants are not statistically significant, and strong models often match or surpass multimodal results using only text—evidence that current VLMs lean heavily on textual priors rather than genuinely vision-grounded reasoning.

\section{Limitations}
While MathSight offers a focused lens on vision-grounded mathematical reasoning, it has several limitations. 
\textbf{Model coverage: }Our experiments primarily evaluate the latest Qwen-series VLMs, with limited exploration of other model families. Broader cross-family evaluation (e.g., OpenAI, Google, Anthropic, Meta, Mistral) is needed to validate the generality of our findings.
\textbf{Dataset scale:} The multimodal portion of the benchmark is modest in size. Larger, more diverse visual sets—covering varied diagram styles, noise conditions, and real-world capture artifacts—would improve statistical power and ecological validity.
\textbf{Benchmark breadth:} Our conclusions should be tested across a wider range of multimodal math benchmarks and related tasks to assess robustness, transferability, and potential domain-specific effects.
\textbf{Task scope:} MathSight focuses on university-level problems; extending to lower grades, professional domains (e.g., engineering drawings), and step-by-step solution formats may reveal different dependencies on visual information.
We view these as practical next steps to strengthen external validity and to better separate linguistic priors from genuinely vision-grounded reasoning.

\bibliography{custom}

\appendix
\onecolumn
\section*{Appendix ------ MathSight}

\section{Comparison of Existing Benchmarks in Text-only and Multimodal} \label{sec:Existing Benchmarks}

\begin{table*}[htbp]
\centering
\caption{Existing Benchmarks, where "Vi.Var" indicates that visual input has different variants and "Pro.Q" stands for proving questions, i.e., questions that cannot be simply verified against a standard answer.}
\label{tab:bench_list}
\begin{adjustbox}{max width=0.9\linewidth}  
\begin{tabular}{llccccccc}
\specialrule{.16em}{0pt}{.65ex}
Benchmark &Venue  &Uni. Level &Other Level &Pure-text & Multimodal & Grad. Level &Pro.Q &Vi.Var\\
\specialrule{.05em}{.4ex}{.65ex}
MathQA \cite{Amini:19MathQA} & ACL'19 & 0 & 37,200 & 37,200 & 0 &\xmark &\xmark &\xmark \\ 
$MMLU_{math}$ \cite{Hendrycks:20MMLU} & ICLR'21 &200 &648 &848 &0 &\xmark &\xmark &\xmark \\
GSM8K \cite{Cobbe:21GSM8K} &arXiv'21 & 0 & 1,319 & 1,319 & 0 & \xmark & \xmark &\xmark \\ 
Math \cite{Hendrycks:21MATH} &NeurIPS'21 & 0 & 5,000 & 5,000 &0 & \xmark & \xmark &\xmark \\ 
Lila \cite{Mishra:22LILA} &EMNLP'22 & 86,742 & 47,073 & 133,815 &0 & \xmark &\xmark &\xmark \\
GSM-Plus \cite{li2024gsm} &ACL'24 & 0 & 10,552 & 10,552 & 0  &\xmark &\xmark  &\xmark \\
MathBench \cite{liu2024mathbench} &ACL Findings'24 & 932 & 2,777 & 3,709  &0 &\xmark &\xmark &\xmark \\
Mamo \cite{huang2024mamo} &arXiv'24 & 557 & 652 & 1,209 & 0 &\xmark &\xmark &\xmark \\ 
PolyMATH \cite{Gupta:24Polymath} &arXiv'24  & 6,750 & 2,250 & 9000 & 0 & \xmark &\xmark &\xmark \\ 
HARDMath \cite{fan2024hardmath} &ICLR'25 & 1,466 & 0 & 1,466 & 0 & \cmark &\xmark &\xmark \\
UGMathBench \cite{xu2025ugmathbench} &ICLR'25  & 5,062 & 0 & 5,062 & 0 & \xmark &\xmark &\xmark \\
\specialrule{.05em}{.4ex}{.65ex}
TheoremQA \cite{chen2023theoremqa} &EMNLP'23 & 800 & 0 & 749 & 51 & \xmark & \xmark &\xmark \\ 

MathVista \cite{lu2023mathvista} &ICLR'24 & 661 & 5,480 & 0 & 6,141 & \xmark &\xmark &\xmark \\ 
Scibench \cite{Wang:23SciBench} &ICML'24 & 869 & 0 & 692 & 177 & \xmark & \xmark &\xmark \\ 
QRData \cite{liu2024qrdata} &ACL Findings'24 & 411 & 0 & 0 & 411 &\xmark &\xmark &\xmark \\ 
MATH-Vision \cite{wang2024math-vision} &NeurIPS'24 & 0 & 3,040 & 0 & 3,040 & \xmark &\xmark &\xmark \\ 
$MMMU_{math}$\cite{Yue:24MMMU} &CVPR'24  & 540 & 0 & 0 & 540 &\xmark &\xmark &\xmark \\ 
U-Math \cite{ko2024umath} &arXiv'24  & 1,100 & 0 & 880 & 220& \xmark &\xmark &\xmark\\ 
Dynamath \cite{Zou:24DynaMath} &ICLR'25  & 1,610 & 3,400 & 310 & 4,700& \xmark &\xmark &\xmark \\ 
MathCheck \cite{zhou2024your} &ICLR'25 & 0 & 4,536 & 3,096 & 1,440 & \xmark &\xmark &\xmark \\ 
\specialrule{.05em}{.4ex}{.65ex}
\textbf{MathSight} (our work) &  NA   & 2048 &0 &1,387 &661 &\cmark &\cmark &\cmark\\
\specialrule{.16em}{0pt} {.65ex}
\end{tabular}
\end{adjustbox}
\end{table*}

\section{Dataset Composition} \label{Dataset Composition}

\subsection{Overview of MathSight}

\begin{figure*}[htbp]
    \centering
    \includegraphics[width=0.9\linewidth]{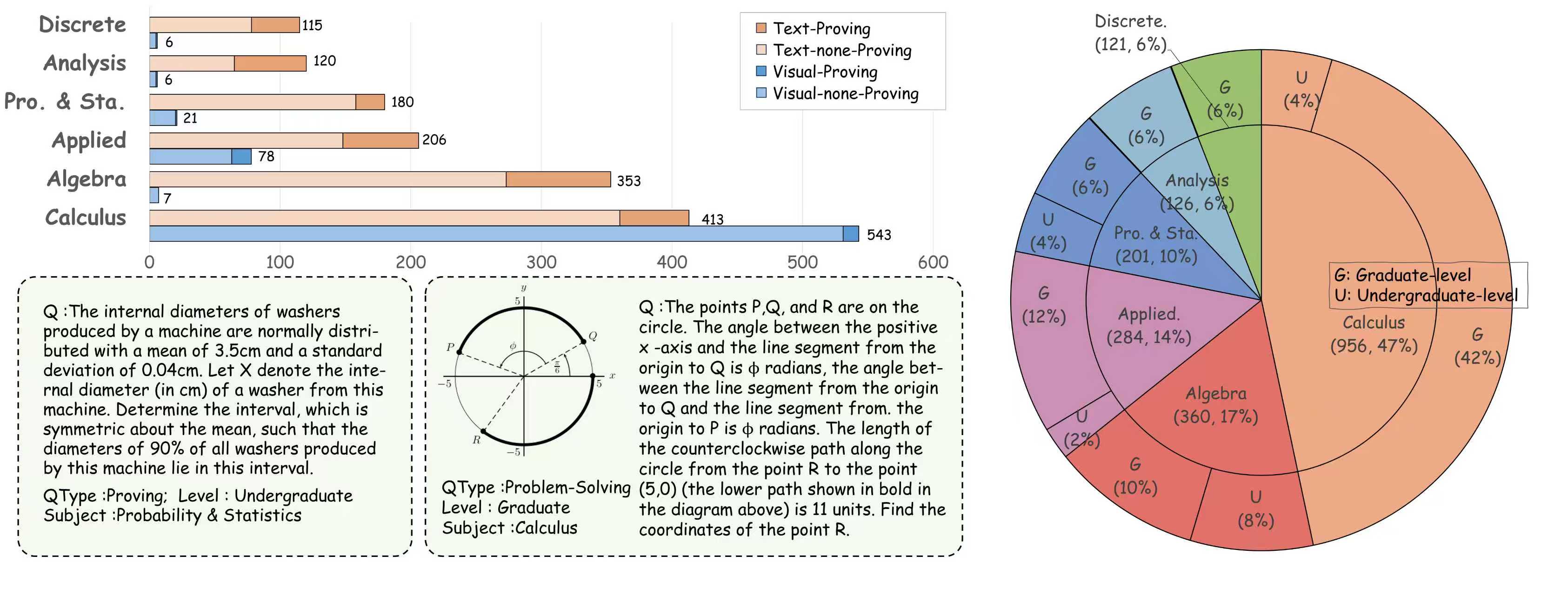}
    \caption{Overview (Categories and examples) of MathSight. 
    }
    \label{fig:data_sta}
\end{figure*}

Figure~\ref{fig:data_sta} is overview (Categories and examples) of MathSight. 

\subsection{The Pipeline of MathSight Construction} 

\begin{figure*}[htbp]
\centering
\includegraphics[width=0.8\linewidth]{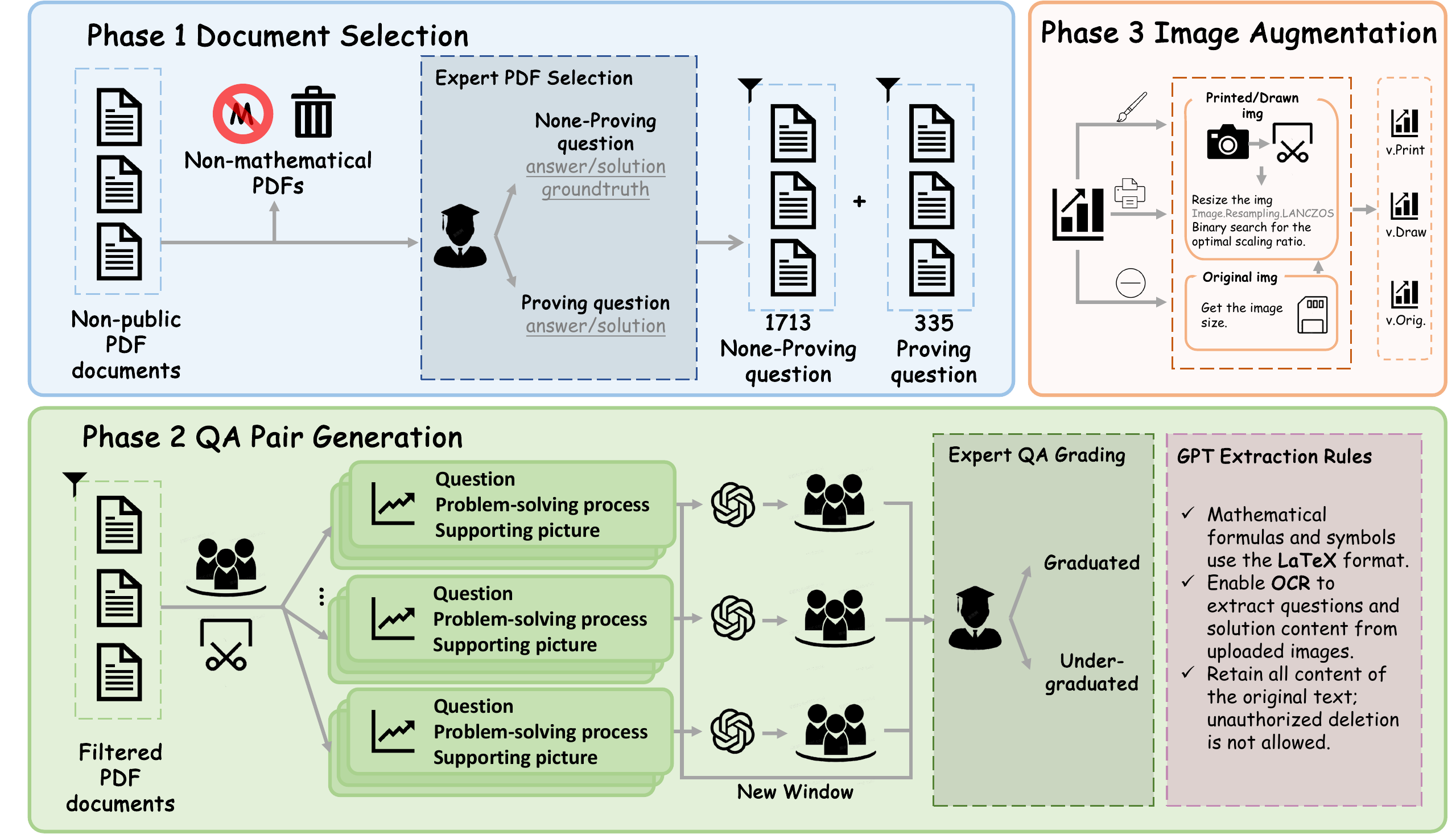} 
\caption{The construction pipeline of MathSight.}
\label{fig:ummbench_pipeline}
\end{figure*}

Figure \ref{fig:ummbench_pipeline} depicts the construction pipeline of MathSight.

 \subsection{Mathematics Problem Extraction Task Processing Template (Reusable)}

A reusable extraction strategy was designed to guide the model, ensuring consistently well-formatted outputs. A detailed and reproducible version of this extraction strategy is presented below.
\onecolumn
\begin{tcolorbox}[colback=white, colframe=gray!75!black,
  title=1.\;General Processing Requirements,
  boxrule=0.5mm, width=0.9\textwidth, arc=3mm, auto outer arc, center]

\begin{enumerate}
  \item \textbf{Output Format.}  
  Output format should be JSONL. Each line must contain the following fields:
  \begin{itemize}
    \item \texttt{question}
    \item \texttt{answer/solution}
    \item \texttt{ground\_truth}
    \item \texttt{image\_path}
  \end{itemize}

  \item \textbf{Preservation of Original Text.}  
  Both the problem statement and the solution must be fully retained in their original form (including derivations, symbols, and formulas). Simplification, summarization, or editing is not allowed.

  \item \textbf{Sub-question Handling.}
  \begin{itemize}
    \item If the problem contains sub-questions (a)(b)(c)\dots, all sub-questions must supplement the full main problem statement.
    \item Remove markers such as (a)/(b) and maintain continuous language.
    \item Each sub-question must keep the entire main problem statement completely intact without any omission.
  \end{itemize}

  \item \textbf{Enable OCR.}  
  Enable image recognition (OCR) to read problem statements and solutions from uploaded images or PDFs.

  \item \textbf{Correct Understanding of \texttt{ground\_truth}.}  
  For non-proof problems with a final computed result (e.g., equals some expression), the \texttt{ground\_truth} field must record this final result. If there is no final result, leave the \texttt{ground\_truth} field blank.  
  If the question asks for the determined answers of multiple parameters, \texttt{ground\_truth} should list each parameter’s final determined answer. For True/False questions, \texttt{ground\_truth} should be marked as \texttt{true} or \texttt{false} accordingly.

  \item \textbf{Mathematical Notation in \mbox{\LaTeX{}}.}
  All mathematical symbols and letters must follow \mbox{\LaTeX{}}~math-reasoning standard formatting.
  For example, subscripts in \mbox{\LaTeX{}} should always be written as \(a_i\) or \(a_{\dots}\).

  \item \textbf{Mandatory \mbox{\LaTeX{}} for Formulas and Symbols.}
  Always use \mbox{\LaTeX{}}~formatting for mathematical formulas and symbols; otherwise, errors may occur.

\end{enumerate}

\end{tcolorbox}

\begin{tcolorbox}[colback=white, colframe=gray!75!black,
  title=2.\;Skipping and Exception Handling Rules,
  boxrule=0.5mm, width=0.9\textwidth, arc=3mm, auto outer arc, center]

\begin{enumerate}
  \item \textbf{Skipping Problems Without Solutions.}  
  If a problem has no solution (the \texttt{solution/answer} field is empty), skip that problem.

  \item \textbf{Recording Skipped Problems.}  
  Record the following information for each skipped problem:
  \begin{itemize}
    \item File name
    \item Problem number
    \item Reason for skipping
  \end{itemize}

  \item \textbf{If Processing Stops Midway.}  
  Indicate:
  \begin{itemize}
    \item Which file and which problem (statement or page) it stopped at
  \end{itemize}

  \item \textbf{Handling Images in Answers or Requests.}  
  For answers or problems requiring images, set \texttt{image\_path} to the image’s file name. If no image exists, leave it blank. If an image is uploaded, set that problem’s \texttt{image\_path} to the image’s name.

  \item \textbf{Preserve Full Original Content.}  
  All original content must be preserved without unauthorized deletion. Words may be added only to maintain fluency.
\end{enumerate}
\end{tcolorbox}

\begin{tcolorbox}[colback=white, colframe=gray!75!black,
  title=3–6.\;Additional Processing Instructions,
  boxrule=0.5mm, width=0.9\textwidth, arc=3mm, auto outer arc, center]

\begin{enumerate}
  \item \textbf{Pre-processing Instructions.}  
  Explain how to paginate the processing and how to maintain high-quality results.  
  Provide a table indicating whether \texttt{ground\_truth} and \texttt{image\_path} are used.  
  Indicate whether multiple sub-questions in one problem can be recognized and split.

  \item \textbf{During Processing.}  
  Process problems in groups of three. After each group is processed, export the results as a JSONL file and make it downloadable.

  \item \textbf{Exporting.}  
  From the first problem onward, export as JSONL. You may export in segments to ensure quality.

  \item \textbf{Additional Requirements for Specific Problems.}  
  (Additional requirements to be filled as needed.)
\end{enumerate}

\end{tcolorbox}






\section{Prompt}\label{sec: appendix_prompt}
\subsection{Generation Prompt}
To generate model responses, we use the following zero-shot prompt:

\onecolumn
\begin{tcolorbox}[colback=white, colframe=gray!75!black, 
title=Zero-shot Prompt, boxrule=0.5mm, width=0.85\textwidth, arc=3mm, auto outer arc,center]

You are a university-level mathematics instructor. Given the following question, provide a complete and rigorous solution. The final answer must be clearly marked as `groundtruth`.

If the question is a proof problem, provide the full proof in the solution section, but leave the groundtruth field empty.

Please strictly follow this output format (use English field names exactly as shown):\\

question:\\
<Original question text>\\
answer/solution:\\
<Your detailed solution>\\
groundtruth:\\
<Final answer or leave empty if proof>

\end{tcolorbox}

\subsection{Evaluation Prompt}
To assess whether the model's answer is correct, we use the following evaluation prompts. Specifically, we have adopted different prompts for proof questions and non-proof questions respectively:
\begin{tcolorbox}[colback=white, colframe=gray!75!black, 
title=Evaluation Prompt for Proof Questions, boxrule=0.5mm, width=0.85\textwidth, arc=3mm, auto outer arc,center]

You are a mathematics instructor. Determine whether the following student's proof is logically valid and correctly proves the statement. Ignore formatting, focus only on mathematical logic.
\\

\textbf{Question:} \{question\} \\
\textbf{Reference Proof:} \{ref\_solution\}\\
\textbf{Student's Proof:} \{student\_solution\}\\
Return only "Yes" if the proof is correct or "No" otherwise.

\end{tcolorbox}





\begin{tcolorbox}[
  colback=white,
  colframe=gray!75!black,
  title=Evaluation Prompt for Non-Proof Questions,
  boxrule=0.5mm,
  width=0.85\textwidth,
  arc=3mm,
  auto outer arc,
  center,
  enhanced,
  sharp corners=all,
  halign=justify,        
  fontupper=\normalsize, 
  before skip=8pt, after skip=8pt
]
    
\vspace{1em}

You are a mathematics evaluator. Check whether the student's final answer and ground\_truth are mathematically equivalent to the reference answer and reference ground\_truth. Ignore formatting differences.

\medskip

\textbf{Question:} \{question\}

\textbf{Reference Final Answer:} \{ref\_answer\}

\textbf{Student's Final Answer:} \{student\_answer\}

\medskip
Return only ``Yes'' if the answer is correct or ``No'' otherwise.
\end{tcolorbox}


\section{Subcategories and Keywords of the \textit{Calculus} Primary Category in MathSight. }\label{sec:Calculus disturbtion}

As shown in Table~\ref{tab:category_res}, the multimodal portion of the MathSight dataset contains 543 problems under the \textit{Calculus} primary category. 
This coarse categorization may lead to a perception of imbalance within the multimodal dataset. 
To address this, we further subdivide \textit{Calculus} into secondary categories according to the chapter organization of \textit{The Princeton Companion to Calculus}. 
This refined classification illustrates the diversity of subjects covered and highlights that several key topics are represented by a notably large number of problems.

\begin{table*}[htbp]
\centering
\small
\renewcommand{\arraystretch}{1.2}
\setlength{\tabcolsep}{6pt}
\begin{tabular}{lll}
\toprule
\textbf{Subcategory } & \textbf{Keyword (} & \textbf{Count} \\
\midrule
\multirow{7}{*}{\textbf{Functions and Graphs}} 
& function & 309 \\
& graph & 74 \\
& inverse & 25 \\
& trigonometric & 19 \\
& sin & 14 \\
& cos & 11 \\
& tan & 8 \\
\midrule
\multirow{4}{*}{\textbf{Limits and Continuity}}
& limit & 23 \\
& limit definition & 12 \\
& approach & 8 \\
& continuity & 4 \\
\midrule
\multirow{4}{*}{\textbf{Derivatives and Rules}}
& derivative & 20 \\
& slope & 13 \\
& rate of change & 7 \\
& tangent & 5 \\
\midrule
\multirow{5}{*}{\textbf{Applications of Derivatives and Optimization}}
& optimization & 26 \\
& max & 14 \\
& min & 12 \\
& motion & 10 \\
& velocity & 7 \\
& acceleration & 5 \\
\midrule
\multirow{3}{*}{\textbf{Exponential and Logarithmic Functions}}
& exponential & 22 \\
& logarithm & 18 \\
& ln & 12 \\
\midrule
\multirow{3}{*}{\textbf{Integrals and Fundamental Theorem of Calculus}}
& integral & 36 \\
& area & 21 \\
& area under & 10 \\
\midrule
\multirow{2}{*}{\textbf{Integration Techniques}}
& substitution & 7 \\
& integration by parts & 6 \\
\midrule
\multirow{2}{*}{\textbf{Series and Taylor Expansion}}
& series & 13 \\
& taylor & 7 \\
\midrule
\multirow{3}{*}{\textbf{Differential Equations}}
& differential equation & 12 \\
& growth & 4 \\
& decay & 3 \\
\bottomrule
\end{tabular}
\caption{Breakdown of Calculus Problems by Subcategory and Keyword. 
Each problem is uniquely assigned to one keyword. 
\textbf{Total:} 34 keywords, 956 problems.}
\label{tab:calculus-subcategories}
\end{table*}

\section{Classification Criterion of Undergraduate- and Graduate-level Problems}\label{sec:classify Undergraduate and graduate level}

In this section, we present our self-defined standard for distinguishing between undergraduate- and graduate-level problems. This standard was developed based on our analysis of the dataset and aims to provide a consistent and reproducible classification framework for difficulty labeling. Specifically, the classification is based on three dimensions: curriculum mapping, knowledge depth, and classification principles.

\subsection{Curriculum Mapping}
The curriculum mapping provides an objective basis for distinguishing between undergraduate and graduate levels. Higher education mathematics courses follow a clear progression in content scope and theoretical depth.

\textbf{Undergraduate Level}  
\textit{Typical Courses}: Linear Algebra, Single/Multivariable Calculus, Probability \& Statistics, Ordinary Differential Equations, Introductory Numerical Analysis.  
\textit{Characteristics}: Course content focuses on fundamental theorems and formula applications. Problems are usually within the scope of standard textbook exercises, with relatively simple mathematical notation and limited derivations.

\textbf{Graduate Level}  
\textit{Typical Courses}: Real Analysis, Complex Analysis, Functional Analysis, Partial Differential Equations, Advanced Probability Theory \& Stochastic Processes, Topology \& Geometry, Optimization Theory, Numerical PDE.  
\textit{Characteristics}: Course content emphasizes the rigor of theoretical systems and the understanding of abstract concepts. Problems often originate from graduate-level textbooks, research papers, or comprehensive exams, with more complex notation systems and multi-layered derivations.

\subsection{Knowledge Depth}
\textbf{Undergraduate Level}: Problems typically involve direct applications of core undergraduate courses, with known formulas directly substituted or solved with a single-step derivation. Reasoning chains are short (1--3 steps), and problems are often closed-ended, with given conditions directly leading to the answer.  
\textbf{Graduate Level}: Problems often require formula derivation, proof of conclusions, or integration of knowledge from multiple disciplines. Reasoning chains are long (more than 3 steps), and problems are often open-ended, with multiple possible solution paths.

\subsection{Classification Principles}
\begin{itemize}
    \item \textbf{Disciplinary Breadth Principle}: If a problem spans two or more advanced mathematical fields (e.g., Real Analysis and Probability, Algebra and Topology), it is likely classified as graduate level.
    \item \textbf{Knowledge Depth Principle}: If a problem involves mathematical concepts not covered at the undergraduate level, it is classified as graduate level.
    \item \textbf{Solution Requirement Principle}: If a problem requires proof of a general proposition rather than computing a specific example, it is likely classified as graduate level.
\end{itemize}

\section{Evaluated Large Models}
We provide more detailed information about the evaluated large models in Table \ref{tab:LM_list}.

\begin{table*}[t]
\centering
\caption{Evaluated Large Models.}
\label{tab:LM_list}
\begin{adjustbox}{max width=0.99\linewidth}
\begin{tabular}{llcccc}
\specialrule{.16em}{0pt}{.65ex}
LLM & Size & Organization & Multi-modal & Open-source \\ 
\midrule
\midrule
Qwen2.5-Math & 72B, 7B & Alibaba & \xmark & \cmark \\ 
Qwen2.5-vl & 7B, 3B & Alibaba & \cmark & \cmark \\ 
Qwen3 & 32B, 14B, 4B, 8B & Alibaba & \xmark & \cmark \\
\specialrule{.05em}{.4ex}{.65ex}
InternVL3 & 78B & OpenGVLab & \cmark & \cmark \\ 
\specialrule{.05em}{.4ex}{.65ex}
DeepSeekV3 (2024/12/26) & 671B & DeepSeek AI & \xmark & \cmark \\ 
DeepSeekR1 (2025/1) & 671B & DeepSeek AI & \xmark & \cmark \\
\specialrule{.05em}{.4ex}{.65ex}
GLM-Z1-Air & 32B & Zhipu AI \& Tsinghua University & \xmark & \cmark \\ 
GLM-4.1V-Thinking & 9B & Zhipu AI \& Tsinghua University & \cmark & \cmark \\ 
\specialrule{.05em}{.4ex}{.65ex}
Kimi-k2-0711-preview & 1T & Moonshot AI & \xmark & \cmark \\ 
\specialrule{.05em}{.4ex}{.65ex}
Claude-3-7-sonnet-20250219 & - & Anthropic & \cmark & \xmark \\ 
Claude-3-7-sonnet-20250219-thinking & - & Anthropic & \cmark & \xmark \\ 
Claude-sonnet-4-20250514 & - & Anthropic & \cmark & \xmark \\ 
Claude-sonnet-4-20250514-thinking & - & Anthropic & \cmark & \xmark \\ 
Claude-opus-4-20250514 & - & Anthropic & \cmark & \xmark \\ 
Claude-opus-4-20250514-thinking & - & Anthropic & \cmark & \xmark \\ 
\specialrule{.05em}{.4ex}{.65ex}
Doubao-seed-1.6-250615 & - & ByteDance & \xmark & \xmark \\ 
Doubao-seed-1.6-thinking-250615 & - & ByteDance & \xmark & \xmark \\ 
Doubao-1.5-thinking-vision-pro-250428 & - & ByteDance & \cmark & \xmark \\ 
\specialrule{.05em}{.4ex}{.65ex}
Gemini-2.5-pro-preview-06-05 & - & Google DeepMind & \cmark & \xmark \\ 
Gemini-2.5-pro-preview-06-05-thinking & - & Google DeepMind & \cmark & \xmark \\ 
\specialrule{.05em}{.4ex}{.65ex}
GLM-4-Plus & - & Zhipu AI \& Tsinghua University & \xmark & \xmark \\ 
GPT-4 & - & OpenAI & \xmark & \xmark \\
GPT-4-turbo (2024/4/9) & - & OpenAI & \xmark & \xmark \\
GPT-4.1 (2025/04/14) & - & OpenAI & \xmark & \xmark \\ 
GPT-4.1-mini (2025/04/14) & - & OpenAI & \xmark & \xmark \\ 
GPT-4o-mini & - & OpenAI & \xmark & \xmark \\ 
GPT-o4-mini (2025/04/16) & - & OpenAI & \xmark & \xmark \\ 
GPT-4o (2024/5/13) & - & OpenAI & \cmark & \xmark \\ 
\specialrule{.16em}{0pt}{.65ex}
\end{tabular}
\end{adjustbox}
\end{table*}

\section{Evaluation Results on Different Question Types} \label{sec: appendix_proving and performaces on different models}
\begin{figure}[t]
    \centering
    \includegraphics[width=0.40\linewidth]{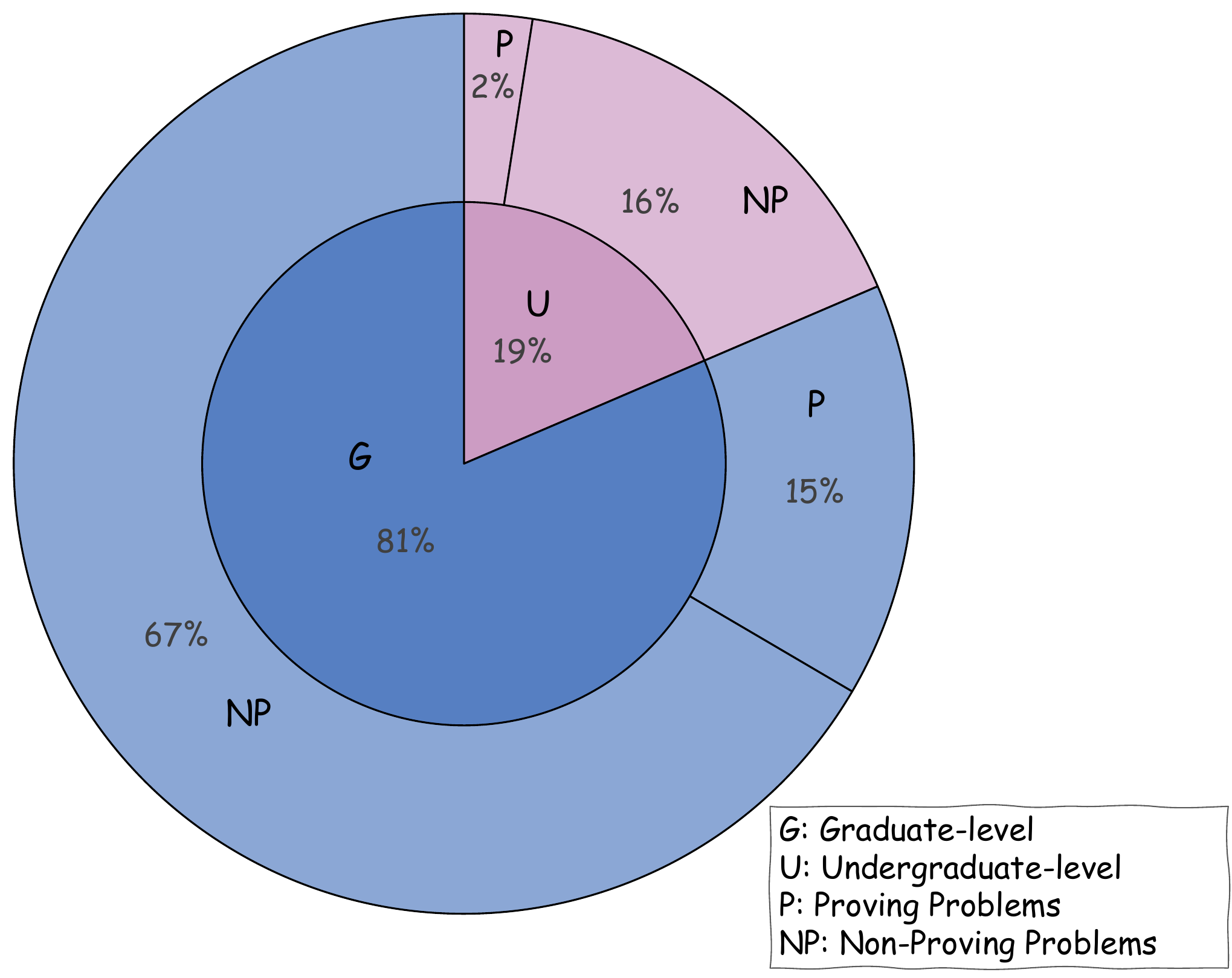}
    \caption{
        The proportion of different question types at the undergraduate and graduate levels.
    }
    \label{fig:proportion of Undergraduate and Proof}
\end{figure}

\begin{figure}[t]
    \centering
    \includegraphics[width=0.95\linewidth]{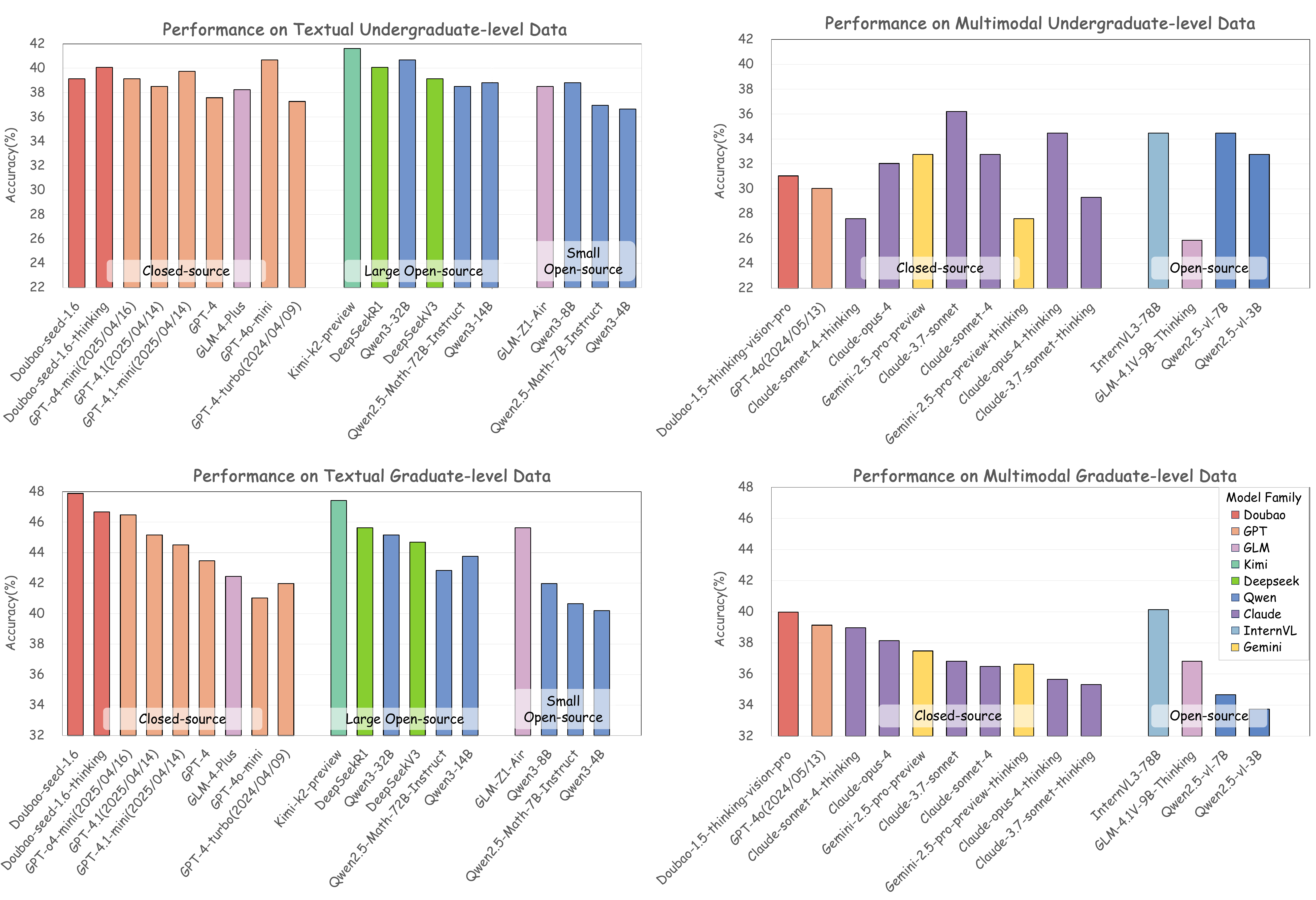}
    \caption{
        The proportion of different question types at the undergraduate and graduate levels.
    }
    \label{fig:diffrent_level}
\end{figure}

\begin{figure*}[t]
\centering
\includegraphics[width=0.9\textwidth]{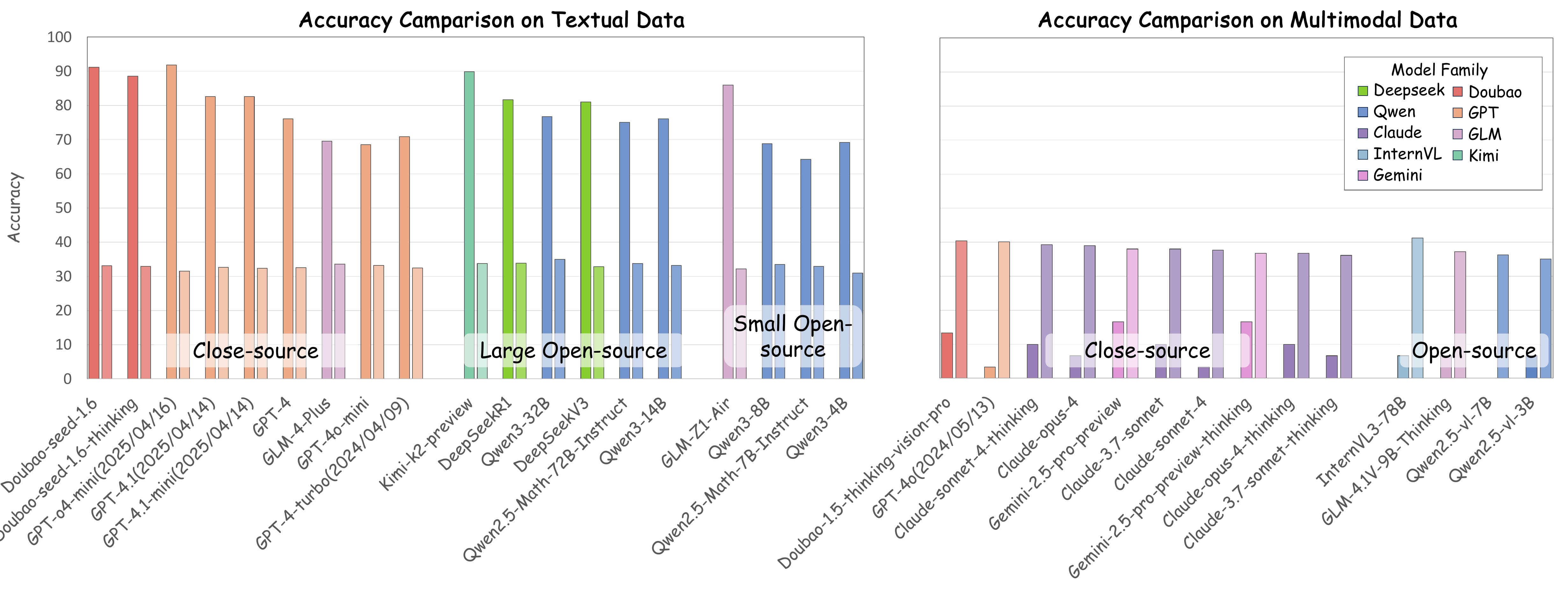} 
\caption{Accuracy comparison of proving and non-proving problems.}
\label{fig:proving_accuracy}
\end{figure*}
We further analyzed the accuracies of different models on proving and non-proving problems. As shown in Figure~\ref{fig:proportion of Undergraduate and Proof}, our problem set contains 380 undergraduate-level problems, among which 50 are proving problems. The graduate-level set contains 1668 problems, with 285 of them being proving problems. Therefore, we can calculate the proportion: proving problems account for 13.16\% of the undergraduate-level problems, while they account for 17.09\% of the graduate-level problems. This difference is one of the reasons why, in Figure~\ref{fig:diffrent_level}, the accuracy of undergraduate-level problems is lower than that of graduate-level problems.

In addition, as shown in Figure~\ref{fig:proving_accuracy}, we calculated two types of accuracies for each model in the problem set:  
(1) the proportion of correctly solved proving problems among all proving problems, which we call \textit{Right-proving accuracy};  
(2) the proportion of correctly solved non-proving problems among all non-proving problems, which we call \textit{Right-non-proving accuracy}.  

For example, for \textbf{Doubao-seed-1.6-250615}, the left bar represents the \textit{Right-proving accuracy} at 91.15\%, while the right bar represents the \textit{Right-non-proving accuracy} at 33.09\%. As shown in the left panel of Figure~\ref{fig:proving_accuracy}, large language models on the textual problem set clearly exhibit a higher \textit{Right-proving accuracy} than \textit{Right-non-proving accuracy}, which explains why in Figure~\ref{fig:diffrent_level} the accuracy for undergraduate-level problems is lower than for graduate-level problems.

Meanwhile, as shown in the right panel of Figure~\ref{fig:proving_accuracy}, for multi-modal models the \textit{Right-proving accuracy} is slightly lower than the \textit{Right-non-proving accuracy}. However, since the multi-modal set contains only 30 proving problems out of 661 problems in total, the proportion is very small. As a result, the difference between \textit{Right-proving accuracy} and \textit{Right-non-proving accuracy} has minimal impact on the undergraduate- and graduate-level accuracies of multi-modal models shown in Figure~\ref{fig:diffrent_level}.

\begin{table*}[tb]
\centering
\caption{Evaluation results of large models on multimodal data of UMMBench under different-sized images. The \textbf{bold} and \underline{underline} numbers represent the best and second-best results of each column, respectively. }
\label{tab:mm_size_res}
\begin{adjustbox}{max width=0.99\linewidth}
\begin{tabular}{llccc}
\specialrule{.16em}{0pt} {.65ex}
\midrule
\multicolumn{4}{c}{\textit{\textbf{Results on Multimodal Data with Different-sized Images}}}\\
\midrule
Model  & $V.Draw_{large}$ & $V.Draw_{small}$ & $V.Photo_{large}$  & $V.Photo_{small}$ \\ 
\specialrule{.05em}{.4ex}{.65ex}
\multicolumn{4}{c}{\textit{\textbf{Closed-source Language Models}}}\\
\hdashline
\midrule
Claude-3.7-sonnet-20250219 & 39.79 & 39.64 & 38.12 & 40.39\\
Claude-3.7-sonnet-20250219-thinking & \textbf{42.36}  & 41.60 & 40.70 & 41.45\\
Claude-sonnet-4-20250514 & 41.00 & 37.67 & 41.45 & 40.70 \\ 
Claude-sonnet-4-20250514-thinking & \textbf{42.36} & 41.60 & 40.7& 41.45\\ 
Claude-opus-4-20250514 & 40.70 & 37.97 & 41.30 & 39.79 \\ 
Claude-opus-4-20250514-thinking & 39.64 & 41.45 & 41.15 & 42.36 \\ 
Gemini-2.5-pro-preview-06-05 & 38.88 & 38.73 & 42.36 & 40.85 \\ 
GPT-4o(2024/05/13) & 37.97 & 36.46 & 41.30 & 37.52\\ 
GPT-5 & \underline{41.60} & 41.75 & \underline{42.66} & 40.85 \\ 
\midrule
\multicolumn{4}{c}{\textit{\textbf{Open-source Multimodal Models}}}\\
\hdashline
\midrule
InternVL3-78B & \textbf{42.51} & \textbf{39.49} & \textbf{42.06} & \textbf{41.30} \\ 
GLM-4.1V-9B-Thinking & \underline{41.60} & 37.37 & \underline{40.24} & 38.58 \\ 
Qwen2.5-vl-7B & 37.97 & 34.95 & 38.58 & 36.31 \\ 
Qwen2.5-vl-3B & 39.49 & 36.61 & 39.94 & 36.61 \\ 
Qwen3-vl & 38.43 & \underline{38.73} & 39.64 & \underline{39.33} \\ 
\specialrule{.16em}{0pt} {.65ex}
\end{tabular}
\end{adjustbox}
\end{table*}


\subsection{The Comparison of Large Models on Undergraduate- and Graduate-level Data}


As shown in Figure~\ref{fig:diffrent_level}, we calculated the accuracies of large models on undergraduate- and graduate-level problems separately. \textit{Counterintuitively}, the performance of large models on graduate-level problems exceeds that on undergraduate-level problems. Upon further analysis, we identified two possible reasons for this result. First, the number of undergraduate-level problems in UMMBench is noticeably smaller than that of graduate-level problems, which may lead to insufficient reflection of the models’ capabilities at the undergraduate level. Second, undergraduate-level problems contains more items in \textit{Algebra} and \textit{Probability $\&$ Statistics}, as shown in Figure~\ref{fig:diffrent_level}. These items need multi-step computations more, which may cause large models to accumulate excessive numerical errors, ultimately leading to a drop in accuracy. These findings may also suggest that large models perform better on proving problems. 

\subsection{Error Analysis}


We conduct error analysis for the textual model \textbf{Doubao-seed-1.6-250615} and the multimodal model \textbf{Doubao-1.5-thinking-vision-pro-250428}. 
From the incorrect problems of each model, we randomly sampled 100 cases for error analysis. 
Then we manually categorized these cases into five predefined error types \cite{xu2025ugmathbench}. 
Specifically, \emph{misunderstanding} refers to incorrect interpretation of the problem; \emph{instruction following error} corresponds to cases where the solution format or required expression was not followed; \emph{numeric calculation error} denotes mistakes in numerical computation or rounding; \emph{express calculation error} indicates structural or symbolic inaccuracies in mathematical expressions; and \emph{partially correct answer} describes solutions with correct intermediate reasoning but incorrect final results.

For \textbf{Doubao-seed-1.6-250615}, the most frequent error type was \emph{partially correct answer}, indicating that the model often established an appropriate problem-solving framework but failed in later derivation or final computation. The second most common error type was \emph{misunderstanding}, suggesting that the model sometimes misinterprets problem statements, especially in complex word problems. Errors related to \emph{numeric calculation} and \emph{express calculation} were relatively rare, implying that once the model correctly interprets the problem, it can usually maintain symbolic and numerical consistency.

For \textbf{Doubao-1.5-thinking-vision-pro-250428}, \emph{misunderstanding} accounted for a significantly higher proportion of errors compared to the text-only model. This is likely due to additional challenges in interpreting visual information or aligning it with textual descriptions. While \emph{express calculation errors} were minimal, \emph{partially correct answer} still represented a large portion of the mistakes, indicating that even when the visual information was processed correctly, reasoning consistency issues persisted.

\begin{figure}[t]
    \centering
    \includegraphics[width=0.80\linewidth]{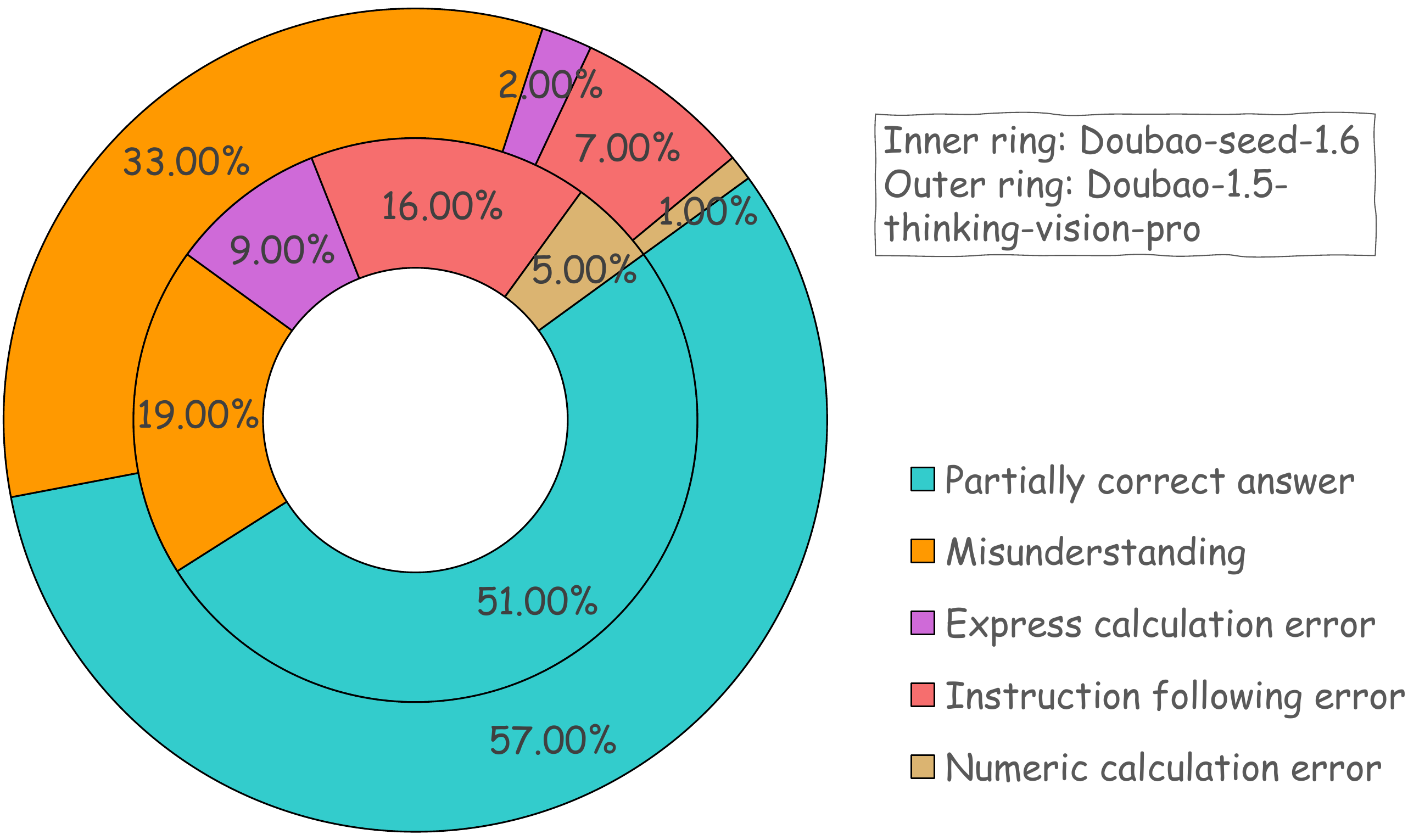}
    \caption{
        Error type distribution of Doubao.
    }
    \label{fig:error-type}
\end{figure}

Overall, both models exhibited a high frequency of \emph{partially correct answer} and \emph{misunderstanding} errors, suggesting that improving problem comprehension and reasoning chain consistency should be priorities for future work. The text-only model tended to make fewer interpretation errors but more expression-related mistakes, whereas the multimodal model struggled more with precise problem understanding, especially in visually grounded tasks. As shown in Figure~\ref{fig:error-type}, partially correct answers and misunderstandings dominate the error distribution for both models, though their proportions differ considerably between the text-only and multimodal settings.

\clearpage
\section{Illustrative Examples of Error Types}

\paragraph{Example 1 (Express calculation error).}
Solve the differential equation:
\[
\frac{dy}{dx} = 2 + \sin(y - 2x),
\]
with initial condition $y(0) = -\frac{\pi}{2}$.

\textbf{Reference Solution:}  
Substitute $z = y - 2x \Rightarrow y = z + 2x$:
\[
\frac{dy}{dx} = \frac{dz}{dx} + 2 = 2 + \sin(z)
\quad\Rightarrow\quad \frac{dz}{dx} = \sin(z).
\]
Integrate:
\[
\int \frac{1}{\sin z} \, dz = \int dx
\quad\Rightarrow\quad \ln\left| \tan\left( \frac{z}{2} \right) \right| = x + C.
\]
Back-substitute $z = y - 2x$:
\[
\ln\left| \tan\left( \frac{y - 2x}{2} \right) \right| = x + C.
\]
Using $y(0) = \frac{\pi}{2}$:
\[
\ln\left( \tan\frac{\pi/4}{1} \right) = C \quad\Rightarrow\quad C = 0.
\]
Final solution:
\[
\ln\left| \tan\left( \frac{y - 2x}{2} \right) \right| = x.
\]

\textbf{Student Solution:}  
Let $v = y - 2x$, then:
\[
\frac{dv}{dx} = \frac{dy}{dx} - 2.
\]
Substitute into the original equation:
\[
\frac{dv}{dx} + 2 = 2 + \sin(v) \quad\Rightarrow\quad \frac{dv}{dx} = \sin(v).
\]
Separate variables and integrate:
\[
\int \csc(v) \, dv = \int dx \quad\Rightarrow\quad \ln|\tan(v/2)| = x + C.
\]
Exponentiating:
\[
\tan\left( \frac{v}{2} \right) = K e^x,\quad K = \pm e^C.
\]
Substitute $v = y - 2x$:
\[
\tan\left( \frac{y - 2x}{2} \right) = K e^x.
\]
Apply $y(0) = -\frac{\pi}{2}$:
\[
\tan\left( -\frac{\pi}{4} \right) = K \quad\Rightarrow\quad K = -1.
\]
Thus:
\[
\tan\left( \frac{y - 2x}{2} \right) = -e^x.
\]
Taking arctangent:
\[
\frac{y - 2x}{2} = -\arctan(e^x)
\quad\Rightarrow\quad y = 2x - 2\arctan(e^x).
\]

\textbf{Error Type:} express calculation error.

\textbf{Comment:}  
While the method and integration were correct, the final expression does not match the reference form. The student’s result introduces an explicit $y$ in terms of $\arctan(e^x)$, whereas the reference keeps the solution in logarithmic–tangent form. This difference in algebraic representation is classified as an express calculation error.


\paragraph{Example 2 (Misunderstanding).}
Estimate the sum $S$ of the series
\[
S = \sum_{n=1}^{\infty} \frac{1}{n^2}
\]
using the first 10 terms (i.e., with $n = 10$).

\textbf{Reference Solution:}  
Using the integral test remainder bounds:
\[
\int_{n+1}^{\infty} \frac{1}{x^2} \, dx \le R_n \le \int_{n}^{\infty} \frac{1}{x^2} \, dx.
\]
So for $n = 10$:
\[
R_{10} \le \int_{10}^{\infty} \frac{1}{x^2} \, dx
= \left[ -\frac{1}{x} \right]_{10}^{\infty} = \frac{1}{10},
\]
\[
R_{10} \ge \int_{11}^{\infty} \frac{1}{x^2} \, dx = \frac{1}{11}.
\]
Thus:
\[
S_{10} + \frac{1}{11} \le S \le S_{10} + \frac{1}{10}.
\]
From numeric calculation:
\[
S_{10} \approx 1.54977 \quad\Rightarrow\quad 1.64068 \le S \le 1.64977.
\]
A good estimate is the midpoint: $S \approx 1.6452$.  
Actually, from Fourier analysis:
\[
\sum_{n=1}^{\infty} \frac{1}{n^2} = \frac{\pi^2}{6} \approx 1.64493.
\]

\textbf{Student Solution:}  
To estimate the sum $S = \sum_{n=1}^{\infty} \frac{1}{n^2}$ using the first 10 terms, we compute the 10th partial sum $S_{10} = \sum_{n=1}^{10} \frac{1}{n^2}$.

Calculating each term and summing them:
\begin{itemize}
\item $n = 1$: $\frac{1}{1^2} = 1.0$ → cumulative sum: $1.0$
\item $n = 2$: $\frac{1}{2^2} = 0.25$ → cumulative sum: $1.25$
\item $n = 3$: $\frac{1}{3^2} \approx 0.1111111$ → cumulative sum: $1.3611111$
\item $n = 4$: $\frac{1}{4^2} = 0.0625$ → cumulative sum: $1.4236111$
\item $n = 5$: $\frac{1}{5^2} = 0.04$ → cumulative sum: $1.4636111$
\item $n = 6$: $\frac{1}{6^2} \approx 0.0277778$ → cumulative sum: $1.4913889$
\item $n = 7$: $\frac{1}{7^2} \approx 0.0204082$ → cumulative sum: $1.5117971$
\item $n = 8$: $\frac{1}{8^2} = 0.015625$ → cumulative sum: $1.5274221$
\item $n = 9$: $\frac{1}{9^2} \approx 0.0123457$ → cumulative sum: $1.5397678$
\item $n = 10$: $\frac{1}{10^2} = 0.01$ → cumulative sum: $1.5497678$
\end{itemize}

Thus, the estimate of $S$ using the first 10 terms is approximately $1.549768$.

\textbf{Error Type:} misunderstanding.

\textbf{Comment:}  
The student computed the partial sum correctly, but misunderstood the task: they were expected to use the integral test remainder bounds to estimate $S$ with an error bound, rather than simply summing the first 10 terms. This demonstrates a misunderstanding of the problem requirements.

\paragraph{Example 3: Partially correct answer.}
Suppose that $X_1, X_2, \ldots$ is a sequence of random variables such that $X_1 \sim \mathrm{Uniform}(0,1)$, and for $n=1,2,\ldots$, the conditional distribution of $X_{n+1}\mid X_1,\ldots,X_n$ is uniform on $[0, c X_n]$ for $c \in (\sqrt{3}, 2)$. Compute $\mathbb{E}[X_n^r]$ for $r>0$.

\noindent\textbf{Reference solution.}\quad
We compute recursively:
\[
\begin{aligned}
\mathbb{E}[X_{n+1}^r]
&= \mathbb{E}\big[\, \mathbb{E}[X_{n+1}^r \mid X_n] \,\big] \\
&= \mathbb{E}\!\left( \frac{1}{c X_n} \int_{0}^{c X_n} y^{r}\,dy \right) \\
&= \mathbb{E}\!\left( \frac{(c X_n)^{r+1}}{(r+1)\,c X_n} \right)
= \frac{c^r}{r+1}\,\mathbb{E}[X_n^r].
\end{aligned}
\]
Recursively applying:
\[
\mathbb{E}[X_n^r]
= \left( \frac{c^r}{r+1} \right)^{n}\, \mathbb{E}[X_1^r]
= \left( \frac{c^r}{r+1} \right)^{n}\cdot \frac{1}{r+1}.
\]

\noindent\textbf{Student solution.}\quad
To compute $\mathbb{E}[X_n^r]$ for $r>0$, we use the law of total expectation and induction.

\emph{Base case ($n=1$).}\;
$X_1 \sim \mathrm{Uniform}(0,1)$, so its density is $f(x)=1$ for $x\in(0,1)$. Then
\[
\mathbb{E}[X_1^r] = \int_0^1 x^r \, dx = \frac{1}{r+1}.
\]

\emph{Recurrence relation ($n\ge 2$).}\;
For $n\ge 2$, $X_n \mid X_{n-1}=x$ is uniform on $[0, cx]$. The conditional expectation is
\[
\begin{aligned}
\mathbb{E}[X_n^r \mid X_{n-1}=x]
&= \int_{0}^{c x} y^r \cdot \frac{1}{c x}\,dy
= \frac{1}{c x}\cdot \frac{(c x)^{r+1}}{r+1}
= \frac{c^r x^r}{r+1}.
\end{aligned}
\]
Hence,
\[
\mathbb{E}[X_n^r]
= \mathbb{E}\!\left[ \frac{c^r}{r+1} X_{n-1}^r \right]
= \frac{c^r}{r+1}\,\mathbb{E}[X_{n-1}^r].
\]

\emph{Inductive proof.}\;
Let $a_n=\mathbb{E}[X_n^r]$. Then
\[
a_n=\frac{c^r}{r+1}a_{n-1},\qquad a_1=\frac{1}{r+1}.
\]
Solving:
\[
\begin{aligned}
&n=1:\; a_1=\tfrac{1}{r+1}, \quad
n=2:\; a_2=\tfrac{c^r}{(r+1)^2}, \quad
n=3:\; a_3=\tfrac{c^{2r}}{(r+1)^3}.
\end{aligned}
\]
By induction, assume $a_k=\dfrac{c^{r(k-1)}}{(r+1)^k}$. Then
\[
a_{k+1}=\frac{c^r}{r+1}\,a_k=\frac{c^{rk}}{(r+1)^{k+1}}.
\]
Thus, for all $n\ge 1$,
\[
\mathbb{E}[X_n^r]=\frac{c^{r(n-1)}}{(r+1)^n}.
\]

\noindent\textbf{Error type.}\; Partially correct answer.

\noindent\textbf{Comment.}\;
The student's recurrence setup is correct, but the final formula
$\frac{c^{r(n-1)}}{(r+1)^n}$ does not match the reference
$\left(\frac{c^r}{r+1}\right)^n \cdot \frac{1}{r+1}$, missing an extra factor $\frac{c^r}{r+1}$ for $n>1$.

\paragraph{Example 4 (Numeric calculation error).}
Find the general solution of the following equation by the method of Variation of Parameters:
\[
y'' - 7y' + 10y = 100x
\]

\textbf{Reference Solution:}  
The homogeneous equation is:
\[
y'' - 7y' + 10y = 0
\]
Characteristic equation:
\[
\lambda^2 - 7\lambda + 10 = (\lambda - 2)(\lambda - 5) 
\quad\Rightarrow\quad \lambda = 2, 5
\]
Fundamental solutions:
\[
y_{1}(x) = e^{2x}, \quad y_{2}(x) = e^{5x}
\]
Compute Wronskian:
\[
W[y_{1}, y_{2}](x) 
= e^{2x} \cdot 5e^{5x} - 2e^{2x} \cdot e^{5x} 
= 3e^{7x}
\]

Let $g(x) = 100x$, apply variation of parameters:
\[
\begin{aligned}
y_{p}(x) &= -e^{5x} \int \frac{100s e^{2s}}{3e^{7s}} \, ds 
+ e^{2x} \int \frac{100s e^{5s}}{3e^{7s}} \, ds \\
&= -\frac{100}{3} e^{5x} \int s e^{-5s} \, ds
+ \frac{100}{3} e^{2x} \int s e^{-2s} \, ds
\end{aligned}
\]
\[
\begin{aligned}
&= -\frac{100}{3} e^{5x} 
\left(-\frac{1}{5}x e^{-5x} - \frac{1}{25}e^{-5x} \right) \\
&\quad + \frac{100}{3} e^{2x} 
\left(-\frac{1}{2}x e^{-2x} - \frac{1}{4}e^{-2x} \right)
\end{aligned}
\]
\[
= \frac{20}{3}x + \frac{4}{3} - \frac{50}{3}x - \frac{25}{3} 
= -10x - 7
\]
General solution:
\[
y(x) = -10x - 7 + c_{1} e^{2x} + c_{2} e^{5x}
\]

\textbf{Student Solution:}  
The student applied the Variation of Parameters method:

1. \;Complementary Solution: Found $y_c = C_1 e^{2x} + C_2 e^{5x}$ from the characteristic roots $2$ and $5$.
2. \;Wronskian: Correctly computed $W = 3e^{7x}$.
3. \;$u_1'$ and $u_2'$:
\[
u_1' = -\frac{e^{5x} \cdot 100x}{3e^{7x}} 
= -\frac{100}{3} x e^{-2x}, 
\quad
u_2' = \frac{e^{2x} \cdot 100x}{3e^{7x}} 
= \frac{100}{3} x e^{-5x}
\]
4. Integration:
\[
u_1 = -\frac{100}{3} \int x e^{-2x} dx
= \frac{50}{3} x e^{-2x} + \frac{25}{3} e^{-2x}
\]
\[
u_2 = \frac{100}{3} \int x e^{-5x} dx
= -\frac{20}{3} x e^{-5x} - \frac{4}{3} e^{-5x}
\]
5. Particular Solution:
\[
\begin{aligned}
y_p &= \left( \frac{50}{3} x e^{-2x} + \frac{25}{3} e^{-2x} \right) e^{2x} \\
&\quad + \left( -\frac{20}{3} x e^{-5x} - \frac{4}{3} e^{-5x} \right) e^{5x}
\end{aligned}
\]
Simplifying:
\[
y_p = \frac{50}{3}x + \frac{25}{3} - \frac{20}{3}x - \frac{4}{3} 
= 10x + 7
\]
6. General Solution:
\[
y = C_1 e^{2x} + C_2 e^{5x} + 10x + 7
\]

\textbf{Error Type:} numeric calculation error.

\textbf{Comment:}  
The student followed the correct Variation of Parameters procedure but made a sign error when combining the two terms in $y_p$. This caused the particular solution to be $10x + 7$ instead of $-10x - 7$, leading to a numeric calculation error in the final answer.

\paragraph{Example 5 (Instruction following error).}
Find all values of $c$ so that $\lambda = 2$ is an eigenvalue of the matrix
\[
A = \begin{bmatrix} 4 & -3 \\ 4 & c \end{bmatrix}.
\]
Clearly circle your answer below:

(i) $c = -3$ only \quad
(ii) $c = -4$ only \quad
(iii) $c = 4$ only \quad
(iv) $c = -6$ only \\
(v) All $c$ except $-4$ \quad
(vi) All $c$ except $-6$ \quad
(vii) All $c$ except $6$

\textbf{Reference Solution:}  
We compute:
\[
\det(A - 2I) =
\det\begin{bmatrix} 2 & -3 \\ 4 & c - 2 \end{bmatrix}
= 2(c - 2) + 12
\]
Setting this equal to $0$:
\[
2(c - 2) + 12 = 0 \quad\Rightarrow\quad 2c - 4 + 12 = 0
\]
\[
2c = -8 \quad\Rightarrow\quad c = -4
\]
Correct choice: (ii).

\textbf{Student Solution:}  
To determine the values of $c$ for which $\lambda = 2$ is an eigenvalue of $A$, we require:
\[
\det(A - \lambda I) = 0.
\]
First, compute:
\[
A - 2I = \begin{bmatrix} 4 - 2 & -3 \\ 4 & c - 2 \end{bmatrix}
= \begin{bmatrix} 2 & -3 \\ 4 & c - 2 \end{bmatrix}.
\]
Determinant:
\[
\det(A - 2I) = (2)(c - 2) - (-3)(4)
= 2(c - 2) + 12
\]
Simplifying:
\[
\det(A - 2I) = 2c - 4 + 12 = 2c + 8
\]
Setting equal to zero:
\[
2c + 8 = 0 \quad\Rightarrow\quad c = -4.
\]
Thus, the only value is $c = -4$.

\textbf{Error Type:} instruction following error.

\textbf{Comment:}  
While the computation of $c$ was correct, the final answer was not provided in the required multiple-choice format (i.e., option (ii) should have been explicitly circled as per instructions). This constitutes an instruction-following error.

\section{Case Study: Comparative Analysis of Answer Combinations across Three Image Versions in GPT\textendash5 and Qwen3\textendash VL\textendash235B\textendash A22B}\label{ABC-three version}

From top to bottom are the original image version, the printed version, and the hand-drawn version of GPT-5 or Qwen3-VL-235B-A22B.

\begin{figure}[t]
\centering
\includegraphics[width=0.9\linewidth]{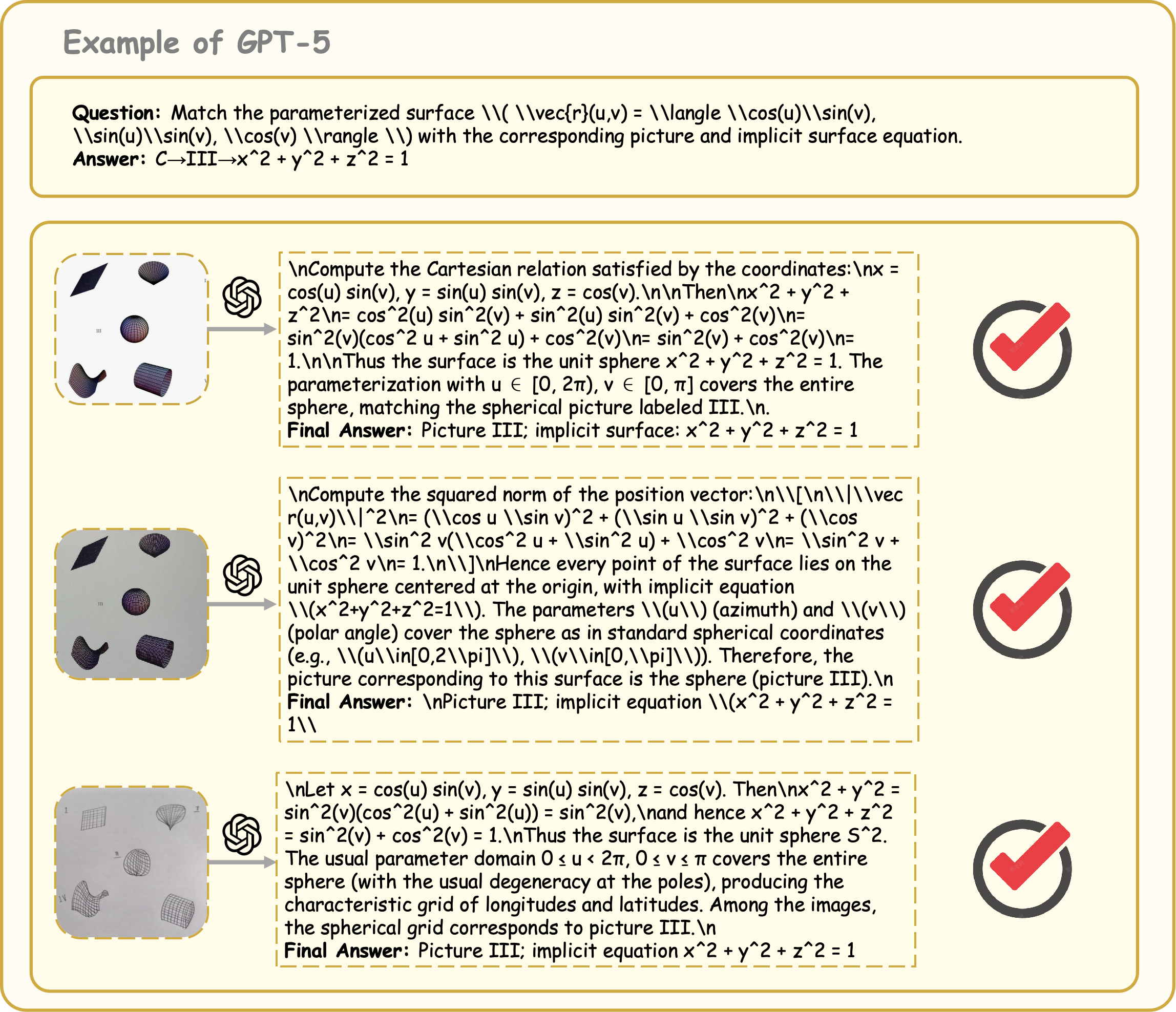}
\caption{GPT-5 Original version is right, printint version is right and drawing version is right}
\label{fig:gpt5-yesyesyes}
\end{figure}

\begin{figure}[t]
\centering
\includegraphics[width=0.9\linewidth]{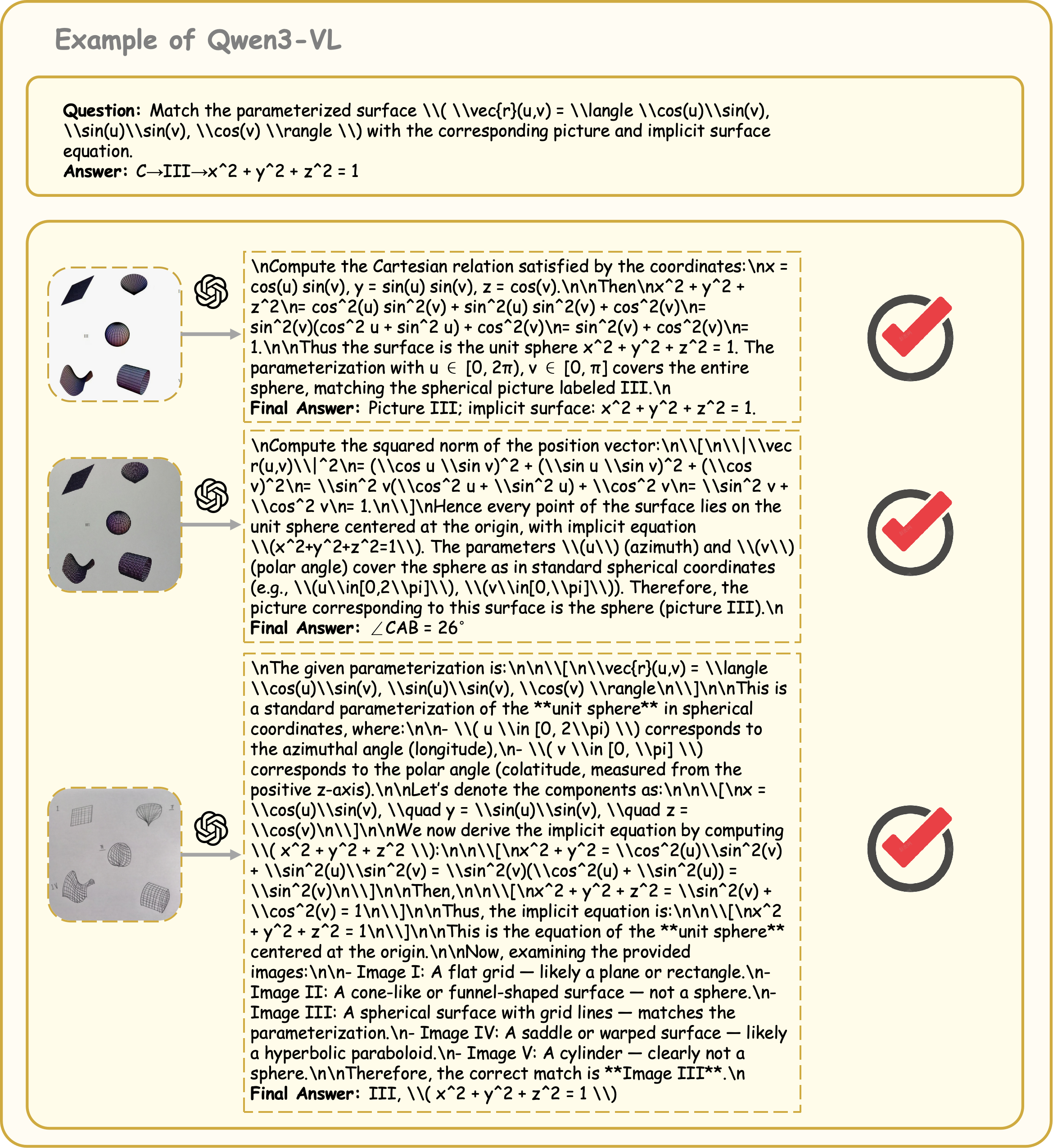}
\caption{Qwen3-VL-A235B-a22B original version is right, printing version is right and drawing version is right}
\label{fig:qwen3-yesyesyes}
\end{figure}

\begin{figure}[t]
\centering
\includegraphics[width=0.9\linewidth]{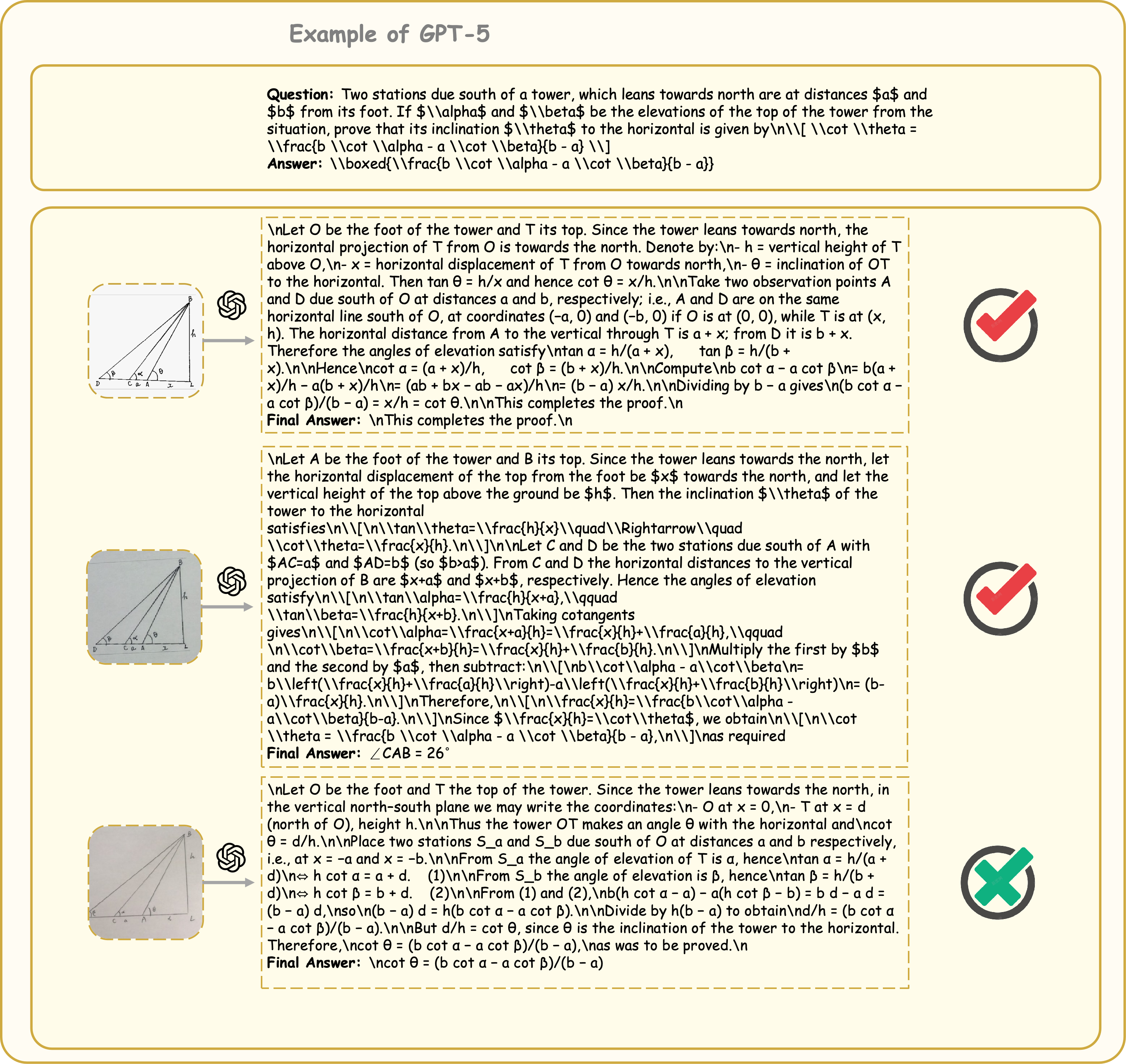}
\caption{GPT-5 original version is right, printing version is right but drawing version is wrong}
\label{fig:gpt5-yesyesno}
\end{figure}

\begin{figure}[t]
\centering
\includegraphics[width=0.9\linewidth]{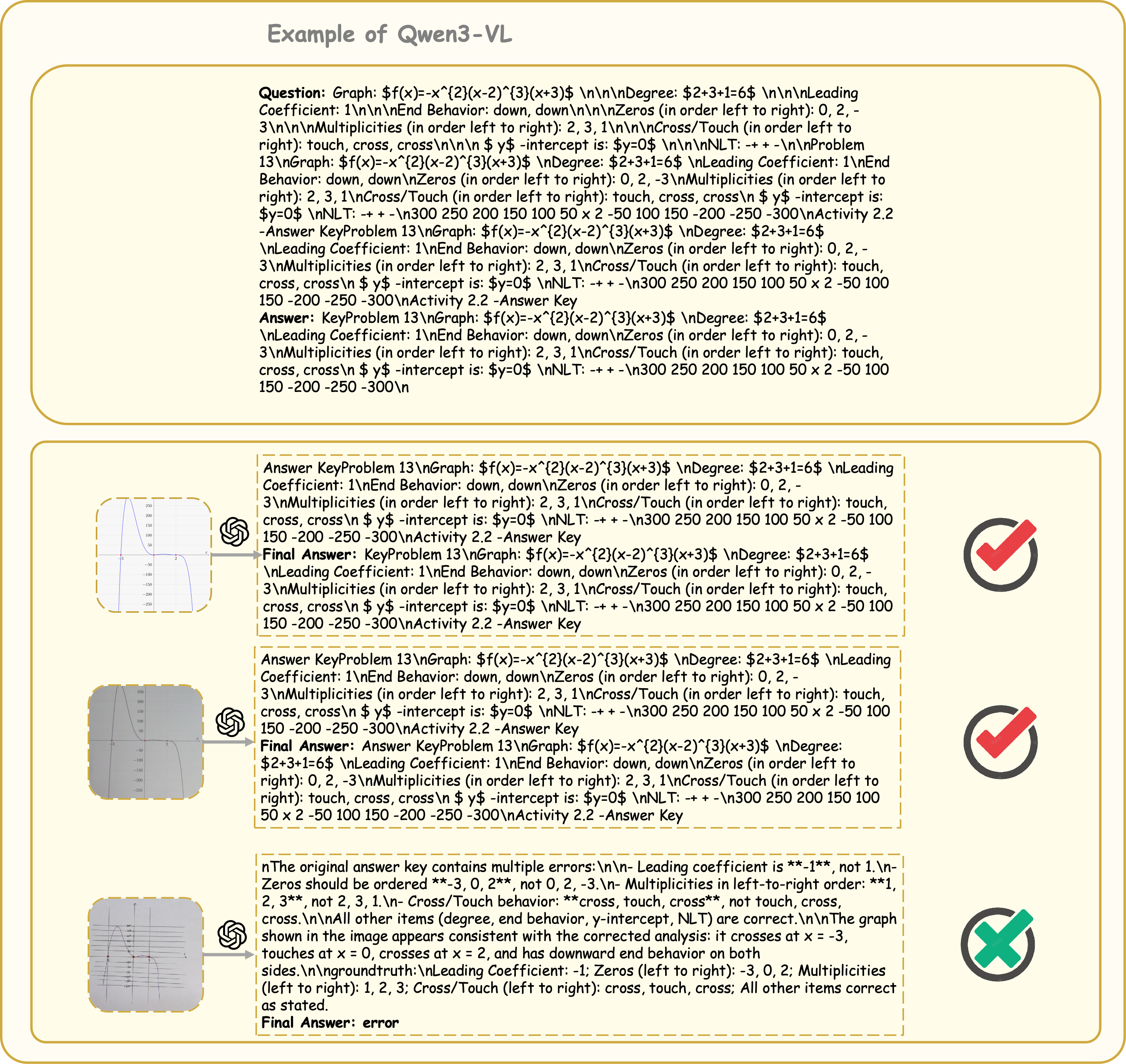}
\caption{Qwen3-VL-A235B-A22B original version is right, printing version is right but drawing version is wrong}
\label{fig:qwen3-yesyesno}
\end{figure}

\begin{figure}[t]
\centering
\includegraphics[width=0.9\linewidth]{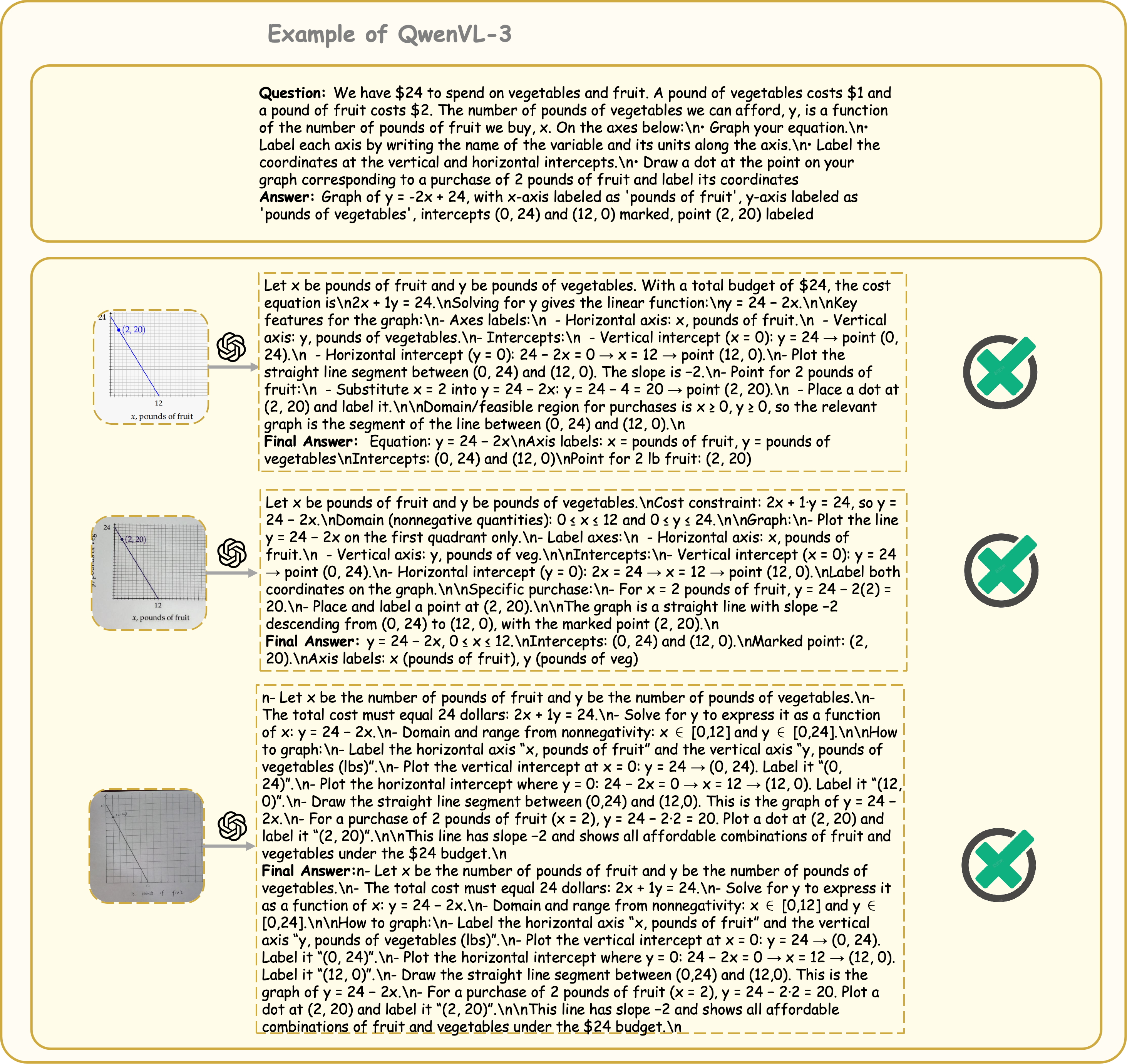}
\caption{Qwen3-VL-A235B-A22B original version is wrong, printing version is wrong and drawing version is wrong}
\label{fig:qwen3-nonono}
\end{figure}

\end{document}